\documentclass{article}

\usepackage[preprint]{corl_2026} 

\usepackage{booktabs}
\usepackage{makecell}
\usepackage{array}
\usepackage{tabularx}
\usepackage[table]{xcolor}
\usepackage{graphicx}
\usepackage{array}
\usepackage{siunitx}
\usepackage{threeparttable}

\newcolumntype{C}{>{\centering\arraybackslash}X}

\usepackage{amssymb} 
\usepackage{pifont}  
\usepackage{bbm}
\usepackage{amsmath}

\usepackage{hyperref}

\usepackage{cleveref}
\usepackage{xspace}

\usepackage{multirow}

\usepackage{titlesec}
\usepackage{algorithm}
\usepackage{algpseudocode}

\titlespacing*{\section}{0pt}{6pt plus 2pt minus 3pt}{3pt plus 1pt}
\titlespacing*{\subsection}{0pt}{5pt plus 1.5pt minus 2pt}{2pt plus 1pt}
\titlespacing*{\paragraph}{0pt}{4pt plus 1pt minus 0.5pt}{0.6em}

\newcommand{\greencheck}{\textcolor{green!60!black}{\checkmark}}
\newcommand{\redcross}{\textcolor{red}{\ding{55}}}
\newcommand{\mediumpfont}{%
  \fontsize{7.5}{9}\selectfont
}

\newcommand{\method}{TaskNPoint\xspace}

\title{\method: How to Teach Your Humanoid\\ to Hit a Backhand in Minutes}

\author{
Blake Werner \quad Ilona Demler \quad Pietro Perona \quad Aaron D. Ames\\
California Institute of Technology\\
\texttt{\{bwerner, idemler, perona, ames\}@caltech.edu}
}

\begin{document}
\maketitle


\begin{figure}[h]
  \centering
  \includegraphics[width=0.95\linewidth]{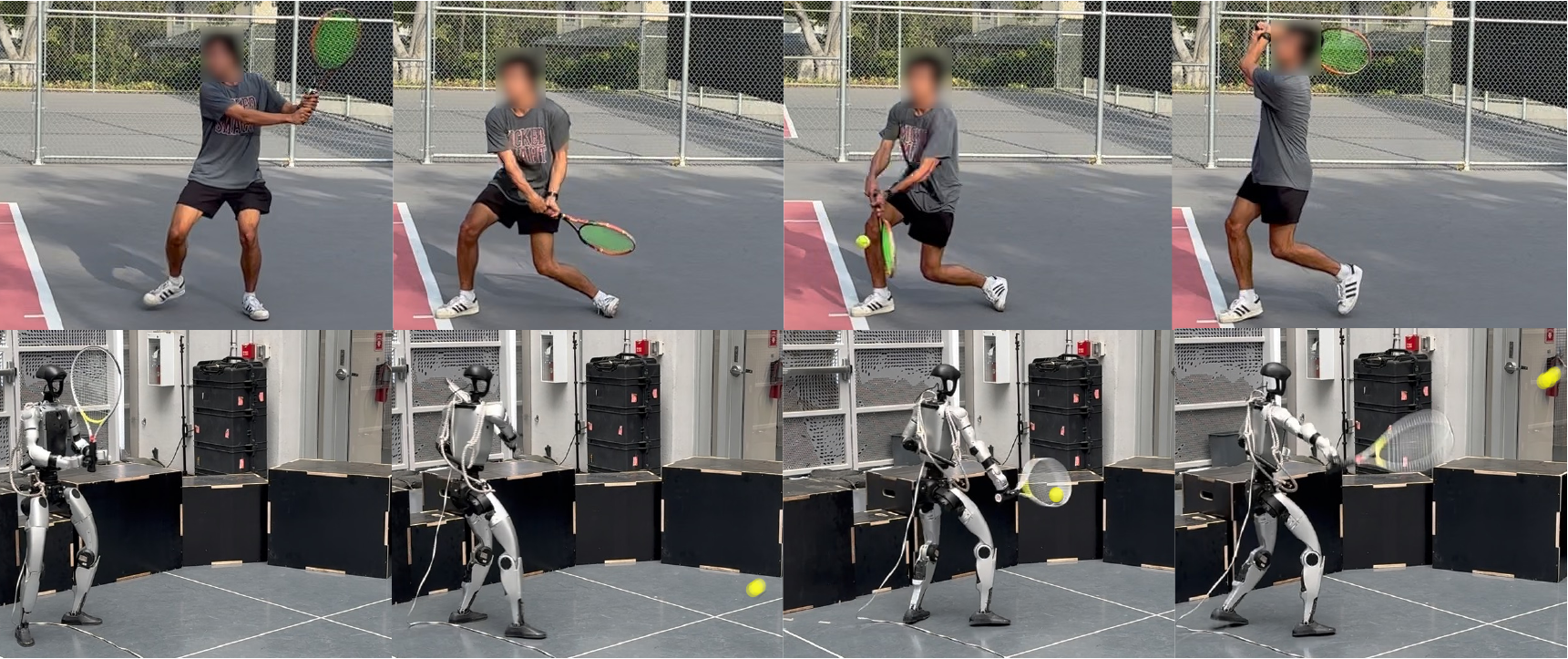}
  \caption{
  \textbf{\method}
  is a training protocol that teaches humanoid robots dynamic skills from a single human video demonstration per motion. We have a human ``coach" specify the discrete skills and critical interaction window, from which we generate goals $G$. Reinforcement learning in simulation with randomized targets lets a single demonstration generalize zero-shot to new target locations, trainable in under an hour on one GPU.
  }
  \label{fig:fig1}
\end{figure}

\begin{abstract} How do we learn to hit a tennis backhand? Not from a thousand hours of tennis tournaments on TV --- we {\em work with a coach} and {\em practice}. We argue this is also the right recipe for teaching dynamic skills to humanoid robots. This follows from a structural property of dynamic skills: the outcome is decided by a short, crucial portion of the trajectory --- for a backhand, the $\sim$20cm of racket travel around ball contact. Getting this {\em interaction window} right requires coordinating the whole motion, so that control, physics, and morphology act in concert. Learning thus reduces to mastering a handful of distinct actions and, for each, practicing until the window comes out right. To this end, we introduce TaskNPoint, a training protocol which makes the coach-learner division of labor explicit. The human coach contributes four inputs: a discrete set of skills (e.g. different shots), one demonstration per skill,  identification of the interaction window, and the goal. Learning in a physically realistic simulation environment fills in each action trajectory and provides robustness to unmodeled events. Crucially, randomized target sampling during training lets a single demonstration generalize zero-shot to unseen goal locations. We test this approach on a Unitree G1 humanoid that hits forehands and backhands against balls thrown by a human, kicks incoming soccer balls, and picks and places boxes from novel locations. We find that learning is successful from short human video demonstrations and under an hour of training on a single GPU, with no per-task reward tuning.
\vspace{0.5em}
\\
\noindent \url{https://ilonadem.github.io/tasknpoint_website/}
\end{abstract}


\section{Introduction}

Suppose you want to learn to play tennis. Probably you will not watch a thousand professionals before you step on the court. Rather, you will work with a coach who will start by defining a shot, e.g. the backhand, demonstrate the motion once or twice, point out what matters (the moment the racket meets the ball), name the goal (get it over the net and inside the lines) and send you to practice. Over the next few sessions you will learn additional shots, each anchored by explanations and a demonstration, and refined through your own practice against varied incoming balls. This is how humans learn sports. We argue that this is also a general strategy to teach humanoid robots dynamic loco-manipulation skills.

Dynamic sports skills have a peculiar structure that makes this kind of learning possible. Although a forehand involves coordinated motion of the entire body, its outcome is determined solely by the position and motion of the racket during contact, or about 20 cm of the racket's trajectory: the {\em interaction window}.  In other words, get the racket head moving in the right direction, at the right speed and time, and the ball will go where you want it. The rest of the swing (the backswing, the footwork, the follow-through) serves the purpose of getting the interaction window right, and is shaped by the morphology of the body and the physics of running, standing, swinging, and balancing under gravity. The same is true for kicking a ball, throwing a javelin, or swinging a bat. 

This observation suggests a natural division of labor between teacher and learner. The teacher contributes expert knowledge: (1) the discrete set of skills that constitute the sport (forehand, backhand, serve, volley, overhead, footwork), (2) a demonstration of each, (3) identification, for each skill, of the  interaction window: the short bit of trajectory that determines success (in striking sports, like tennis, this is the moment of contact, in throwing sports, like bocce, it is the moment of release), and (4) a specification of the goal --- for tennis, placing the ball at a chosen location in the opponent's court, though in this paper we adopt the beginner's goal of racket-ball interaction quality. The learner, via practice in a physics-accurate simulation environment, supplies everything else: filling in the rest of the trajectory, generalizing the skill to a continuum of possible target points and velocities, and acquiring robustness to perturbation through systematic variation of the ball's incoming trajectory. We call this the TaskNPoint paradigm.

We test our paradigm by training dynamic humanoid motions from a handful of monocular videos of tennis, soccer, and box pick-and-place. We find that from one demonstration per skill (a few seconds of video), we can train a Unitree G1 humanoid to hit tennis forehands and backhands, kick soccer balls toward targets, and pick up boxes in new locations. This takes less than an hour on one GPU. Training in simulation achieves sufficient precision and robustness to enable successful hardware deployment with no per-task reward tuning.

\section{Related Work}
\label{sec:related_work}

\begin{figure}[]
  \centering
  \includegraphics[width=\linewidth]{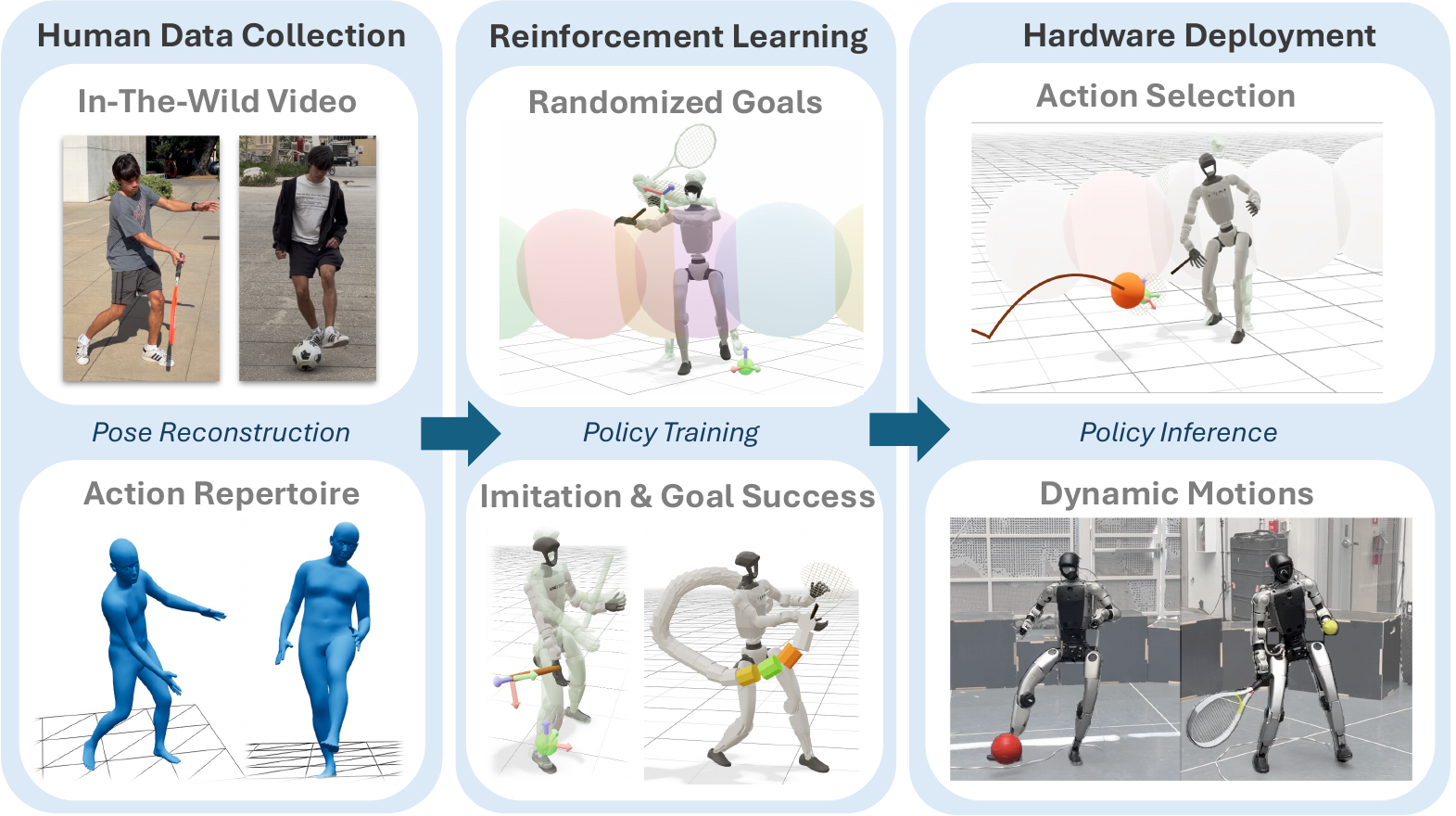}
  \caption{
  \textbf{\method Overview.}
  From a small collection of videos of human demonstrations (one video per task), we provide a pipeline for learning a repertoire of goal-conditioned motions interacting with dynamic environments. We first reconstruct the human motions using SMPLX parameters (\cref{sec:motion_recon}), which we then kinematically retarget to the humanoid (\cref{sec:kinematic_retargeting}). A higher level planner conditions action selection on goal, then a policy trained to both imitate the action and accomplish the goal (\cref{sec:policy_learning}) performs the task conditioned on the simple abstraction, allowing for robust performance during deployment (\cref{sec:result}).
  }
  \label{fig:method}
\end{figure}

\newcolumntype{C}{>{\centering\arraybackslash}X}

\begin{table}[h]
\centering
\caption{\textbf{Comparison to existing methods.}  \method is a general framework for training dynamic humanoid motions. 
Despite being multi-task and trained on dynamic environments, it is faster and more data efficient to learn than competing approaches: for each task it requires just 1 demonstration with seconds of data, and has a simple single-stage training setup that takes less than 1 GPU hour to train to deployable performance. 
For GPU training hours, we assume convergence in 30,000 iterations of RL training at 1 second per iteration, the standard in \cite{makoviychuk2021isaac}, for each policy in the pipeline.
}
\label{tab:comparison_to_existing}

\scriptsize
\setlength{\tabcolsep}{3.5pt}
\renewcommand{\arraystretch}{1.15}

\begin{tabularx}{0.95\textwidth}{@{} l C C C C C C @{}}
\toprule
 & Dynamic Environment
 & Training Stages
 & GPU Training Hours
 & Multi-Task?
 & Train demos per task
 & Imitation Data (min) per task
\\

\midrule

\textbf{VideoMimic \cite{allshire2025visual}}
    & \redcross
    & 4
    & 33
    & \greencheck
    & 17.6
    & 20.5
\\

\textbf{HDMI \cite{weng2025hdmi}}
    & \redcross
    & 1
    & 8
    & \redcross
    & 1.0
    & 1.0
\\

\textbf{OmniRetarget \cite{yang2025omniretarget}}
    & \redcross
    & 1
    & 8
    & \redcross
    & 1.0
    & 0.1
\\

\textbf{LATENT \cite{zhang2026learning}}
    & \greencheck
    & 3
    & 25
    & \redcross
    & $>$200.0
    & 300.0
\\

\textbf{HITTER \cite{su2025hitter}}
    & \greencheck
    & 1
    & 8
    & \redcross
    & 2.0
    & 0.3
\\

\textbf{HumanX \cite{wang2026humanx}}
    & \greencheck
    & 3
    & $>$50
    & \greencheck
    & 1.0
    & 500.0
\\

\rowcolor{gray!15}
\textbf{\method (Ours)}
    & \greencheck
    & 1
    & 1
    & \greencheck
    & 1.0
    & 0.5
\\

\bottomrule
\end{tabularx}
\end{table}

\paragraph{Humanoid Loco-manipulation}
The humanoid form makes human demonstrations a natural prior for loco-manipulation, and the field has utilized this fact in several ways. RL for motion tracking, pioneered by works such as \cite{peng2021amp,peng2018deepmimic}, show excellent results at tracking specific trajectories robustly \cite{liao2025beyondmimic, weng2025hdmi, yang2025omniretarget}, however generalization outside of the imitation data proves to be difficult. Addressing this problem has largely been approached in two different ways. The first optimizes a reward through massive simulations \cite{makoviychuk2021isaac, hwangbo2018per,rudin2022learning, haarnoja2024learning, ma2025learning, wang2025beamdojo}, while the second uses general whole-body controllers that track arbitrary trajectories \cite{wang2026humanx, wu2026perceptive, liao2025beyondmimic, ze2025twist2,yu2021human, luo2024universal} and relies on a higher-level trajectory generator. The first is much more data efficient, using minimal numbers of human demonstrations, but often results in behaviors that are unnatural and inefficient, due to the bias of tuned rewards. The second bypasses reward engineering thanks to learning from massive human motion datasets. This scale is becoming more achievable thanks to improvements in video reconstruction \cite{peng2021neural, moon2024expressive, rajasegaran2022tracking, luvizon20182d, rajasegaran2023benefits, kanazawa2018hmr, kanazawa2019hmmr, wang2025prompthmrpromptablehumanmesh, daniilidis2024tram, xiaowei2024gvhmr, yuan2025genmo, allshire2025visual} and imitation. However, these methods often require expensive ad-hoc data cleaning to ensure high-enough quality for effective training \cite{yuan2021simpoe,yuan2023learning,ugrinovic2024multiphys,zhang2021learning,li2022d}. Our method differs from both of these approaches in two ways. First, we use a principled approach to human reconstruction using MLE estimates from multiple views, which removes the need for reward-engineering or task-specific fixes to imperfect reconstructions. Second, by formulating dynamic tasks as goals conditioned on points in 3D space and time, we are able to train on a small set of these demonstrations and achieve motion coverage over a 3D volume around a player.

\paragraph{Sport-playing Humanoids}
Sport-playing humanoids \cite{wang2025hierarchical,ren2025humanoid,luo2024smplolympics,liu2018learning,liu2025humanoid,kim2025physicsfc,chen2026learning,su2025hitter} demonstrate impressive behaviors, but these policies are often tuned to their sport, and some sidestep whole-body coordination by decoupling upper and lower body \cite{su2025hitter}. Neither matches the data efficiency of a human learner: a tennis student does not invent "forehand" from raw experience \citep{calinon2007teacher,nehaniv2002correspondence}, because coaches have spent centuries cataloging the skills of the game and what matters about each. 
In \method we propose a new paradigm for understanding dynamic tasks as spatio-temporal goals, and use this paradigm  to avoid priors induced by task-specific reward tuning. In doing so, we simplify the learning problem by considering the sport divided into clear, discrete actions that can be trained efficiently in parallel simulation, and results in multi-skill policies that train in under an hour on a single GPU.

\section{Task Definition}

\paragraph{Intuition.} 
 
Consider what a tennis player must do when returning an incoming ball. First, they must decide which shot to play. Is it a forehand, a backhand, a smash, or a volley? This is a discrete choice with a finite vocabulary of actions. 
Second, once the shot is chosen, they must shape the shot 
depending on the incoming ball's speed, spin, and trajectory, and on the desired ball destination. Shaping the shot includes the footwork to reach a good position, and the preparation and follow-through that produce the desired impact quality. This is a choice within a continuum of possibilities. 

We therefore arrive at a natural hierarchy of discrete action selection and continuous action shaping needed to execute the task. The policies for shot selection and for shot shaping need to be specified -- we will use RL for shot shaping (\cref{sec:actions}) and a simple hand-made policy for shot selection (\cref{sec:estimation}) . 
While we use tennis as a running example, the paradigm is quite general. This decomposition similarly applies to many robot loco-manipulation tasks, including those that require multiple sequential targets. Besides tennis, we experiment with ball kicking and box pick-and-place to validate this point.

Regarding the action (the {\em shot}, in tennis), we need to distinguish between four different and interrelated entities: the control, the trajectory, the interaction, and the goal. There is a causal connection between them: the {\em control} will make the robot move and describe a {\em trajectory} --- a time sequence $q_{t=0:T}$, where $q$ indicates the degrees of freedom of the robot ---  defining both the kinematics and the dynamics of the robot during the action. The position and velocity of the end-effector at the time of {\em interaction} determine the outcome which, hopefully, meets the {\em goal}. E.g. in tennis the position and velocity of the racket when it impacts the ball (the interaction) determine the trajectory of the ball, and eventually the placement of the ball into the opponent's court (the goal). Thus, the goal $G$ is a function of the interaction $J$, which is a function of the action's trajectory  $A$, which is a function of the control signal. To select the best control, the one that best places the ball, we must invert this chain. Thus, learning must produce a map, or policy, that computes the control from the goal. Or, in a more structured approach, three concatenated maps: from goal to interaction, from interaction to trajectory, and from trajectory to control.

In this paper we simplify this rather complex picture for the sake of studying the overall \method framework. First, we do not learn the map from control to trajectory since this may be obtained analytically using classical control theory, and is available as a primitive through the humanoid's standard software. Second, like beginners in tennis, we adopt the quality of the impact as the goal, rather than having ball placement as the goal and inverting a target landing location into the impact parameters (which is left for future work). Since we identify $G:=J$, only one map, $G \rightarrow A$, needs to be learned.

\paragraph{Formalization.} Given a reference action trajectory 
$A := q_{t=0:T}$, where $q$ is the vector indicating the degrees of freedom of the robot, the teacher identifies the crucial moment in the demonstration sequence, i.e. the time where the racket impacts the ball in tennis, the foot kicks the ball in soccer, and the hand grabs the box in box pickup.  For a tennis shot the racket-ball interaction is parametrized by the time $t^* \in \mathbb{R}$, the 3D location $p^* \in \mathbb{R}^3$, the velocity direction $\nu^* \in \mathbb{R}^3$ and orientation $n^* \in \mathbb{S}^2$ of the racket at impact. Thus, the parameterization of the interaction in the reference demonstration is $G^* := (p^*,\nu^*,n^*, t^*)$ (similar parameterizations for other tasks). In this paper, this is denoted the {\em goal}.  The rest of the robot's trajectory will be determined taking into account the kinematics and dynamics of the robot, as well as other terms in the loss function, such as torque minimization. 

The teacher provides one, or at most few, demonstrations, which are sufficient to provide a qualitative idea of the trajectory of the shot. It is, however, not sufficient to learn how the interaction (racket-to-ball impact) can achieve a multitude of different impact types (the goal here) and deal with a multitude of incoming ball trajectories. To train a policy that is robust to uncertainty on where the racket will meet the ball (both are traveling at high speed, thus such uncertainty is unavoidable), we generate goals $\mathbb{G}_{\text{train}} = \{ (p,\nu,n, t) \}$, where the impact variables are randomized: $p \sim \mathcal{N}(p^*, \Sigma)$, and similarly for $\nu$ and $n$. To average over noisy sensor measurements we average position, velocity direction, and orientation rewards in training over the interaction window, i.e. a small time-window around impact: $\Omega = t^{*}+(-\delta t,\delta t)$. 
The set of all actions $A$ and goals $G^*$ constitute a library of motions $\mathbb{A}$, which we use to train our policy. 

In the remainder of this paper we will discuss how we can apply this dynamic task abstraction as shown in \cref{fig:method}. We first extract goals $\mathbb{G}$ from human demonstrations (\cref{sec:motions}) which we then use to train a humanoid policy (\cref{sec:policy_learning}). We then define a simple motion-selection optimization for task space coverage and deploy our policy on hardware (\cref{sec:estimation}). Finally, we discuss our results and compare \method to current state-of-the-art methods (\cref{sec:result}).

\section{Motion Acquisition}
\label{sec:motions}

\begin{figure}[]
  \centering
  \includegraphics[width=\linewidth]{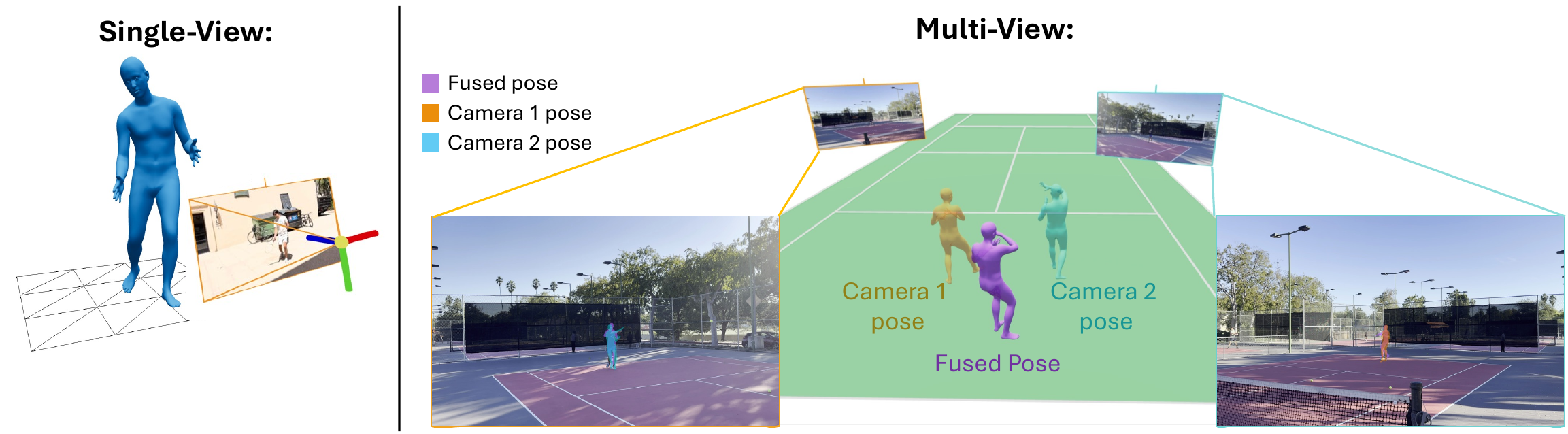}
  \caption{
  \textbf{Video Demonstrations.} We collect single-view (left) and multi-view (right) demonstrations and reconstruct human poses using state-of-the-art reconstruction methods (\cref{sec:motion_recon}). For the multi-view demonstrations, we fuse per-view estimates into a maximum-likelihood pose estimate.}
  \label{fig:motion_recon}
\end{figure}

In this section we outline our method for producing a repertoire of reference motions from a small set of videos of human demonstrations (\cref{fig:motion_recon}).

\paragraph{Motion Reconstruction}
\label{sec:motion_recon}

We start by collecting monocular videos of human demonstrations and extracting human pose estimates using state-of-the-art off-the-shelf monocular human pose estimation and SfM methods (Appendix \ref{app:hpe_recon}). We recover 3D human pose estimates with PromptHMR~\cite{wang2025prompthmrpromptablehumanmesh}, producing per-frame pose estimates of 3D SMPL-X~\cite{pavlakos2019smplx} parameters: $H = \smash{\{(d_t, \phi_t, \beta_t, \theta_t)\}_{t=0}^{T}}$, with translation $d_t \in \mathbb{R}^{3}$, body pose $\theta_t \in \mathbb{R}^{21 \times 3}$, orientation $\phi_t \in \mathbb{R}^{3}$, and shape $\beta_t \in \mathbb{R}^{10}$. 
To counter known failure modes such as foot skating and translation drift~\cite{yeung2025athletepose3d}, for in-the-wild videos we use multiple concurrent views of tennis practices from \cite{demler2026caltennis} and form a maximum-likelihood consensus pose by fusing per-view pose estimates. 
Each per-camera joint estimate $J_c^{(i)}$ is modeled as a Gaussian observation of the true joint position, with covariance $\Sigma_c^{(i)}$, elongated along that camera's depth axis to take into account monocular depth ambiguity. We lift camera-coordinate estimates into a shared world-coordinate frame $J_c \to J_w$, $\Sigma_c \to \Sigma_w$ defined relative to the court geometry (as outlined in Appendix \ref{app:mle_pose}). To fuse multiple estimates into a consensus we compute the maximum-likelihood estimate of the joint positions:
\begin{equation}
    P_{\text{MLE}} = \left( \sum_{i=1}^{N} (\Sigma_w^{(i)})^{-1} \right)^{-1} \sum_{i=1}^{N} (\Sigma_w^{(i)})^{-1} J_{w}^{(i)}.
\end{equation}
We provide the full derivation of the MLE pose in Appendix~\ref{app:mle_pose}.

\paragraph{Kinematic Retargeting}
\label{sec:kinematic_retargeting}

We next retarget human motion trajectories to the robot's morphology and identify the goal $G^* = \{ (p,\nu^*,n^*, t^*) \}$ that will serve as the nominal goal of the motion. First, we kinematically retarget the human motion trajectory with GMR \cite{araujo2025retargeting} to translate the trajectory of the body of the human coach demonstrating a shot, e.g. a tennis backhand, to the closest motion that the robot can execute. A crucial aspect of retargeting is identifying the moment where the interaction happens. Upon identifying the timestamp $t^*$ of interaction, we set the position  of the end effector (e.g. the hand of the tennis player) at $t^*$ as $p^*$, and set the link's velocity direction and orientation as $\nu^*$ and $n^*$. During training the target points $p$ will be randomized around the nominal target $p^*$ with covariance $\Sigma = \text{diag}(\sigma_x, \sigma_y, \sigma_z)$ to cover the 3D volume of possible target locations around the humanoid with sufficiently fine sampling.  We denote the set of all retargeted demonstrations as an action library $\mathbb{A}$.

\section{Policy Learning}
\label{sec:policy_learning}

\begin{figure}[]
  \centering
  \includegraphics[width=1.0\linewidth]{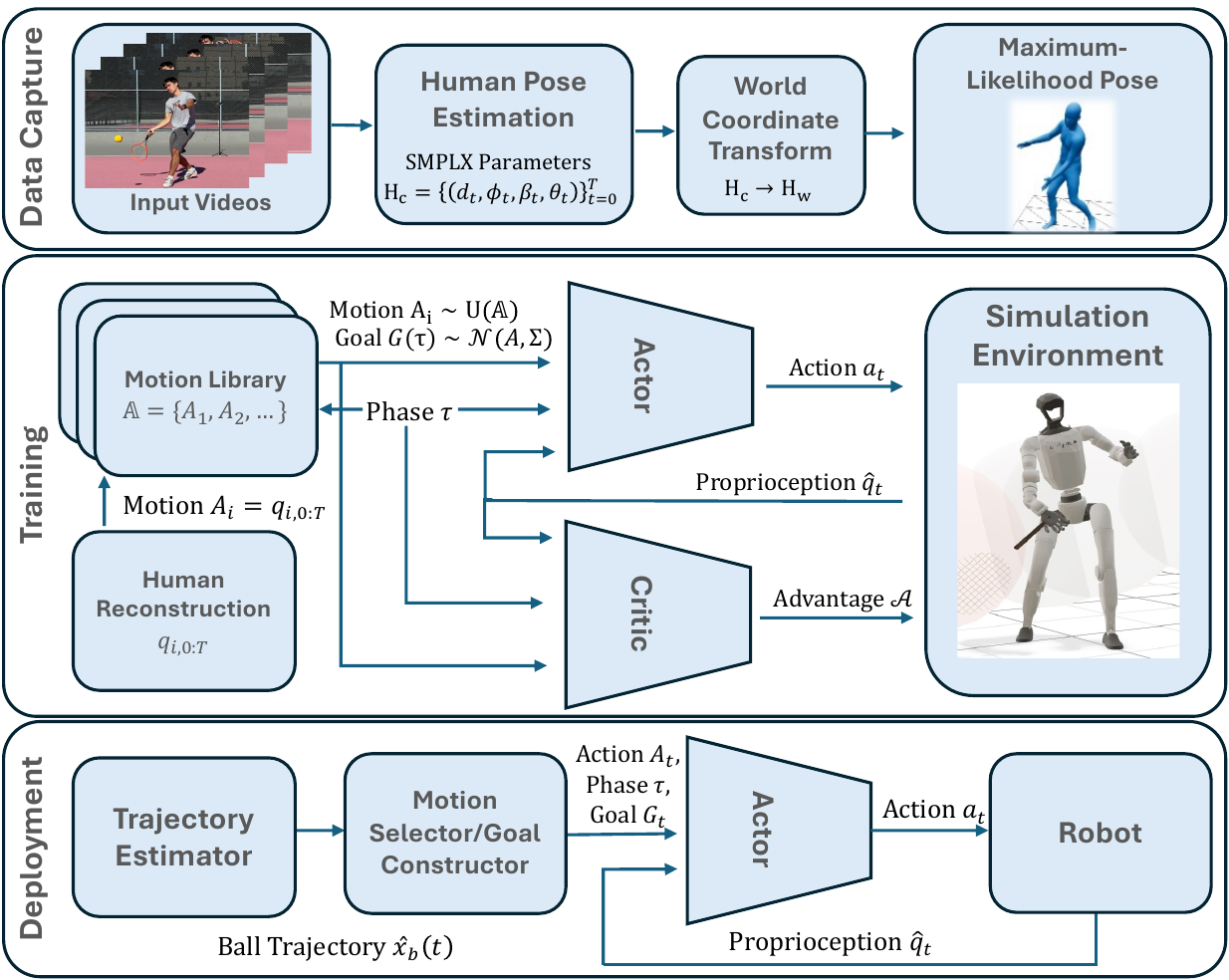}
  \caption{
  \textbf{Policy Architecture.} \textbf{(Top):} We reconstruct reference single-view or multi-view human demonstrations via SMPL-X parameters (\cref{sec:motions}). For multi-view video, we lift reconstructions into a shared coordinate space and calculate a maximum likelihood pose. \textbf{(Middle):} In training we use Asymmetric Actor Critic Policy Optimization \cite{schulman2017proximal} to optimize motions $A_i$ from our motion library $\mathbb{A}$, from which we sample randomized goals $G$ (\cref{sec:commands}). \textbf{(Bottom):} In deployment, policy observations are computed by a trajectory estimator and fed to the motion selection algorithm which takes care of selecting the action type, goal, and phase (\cref{sec:estimation}).}
  \label{fig:learning_architecture}
\end{figure}

\begin{figure}[]
  \centering
  \includegraphics[width=0.9\linewidth]{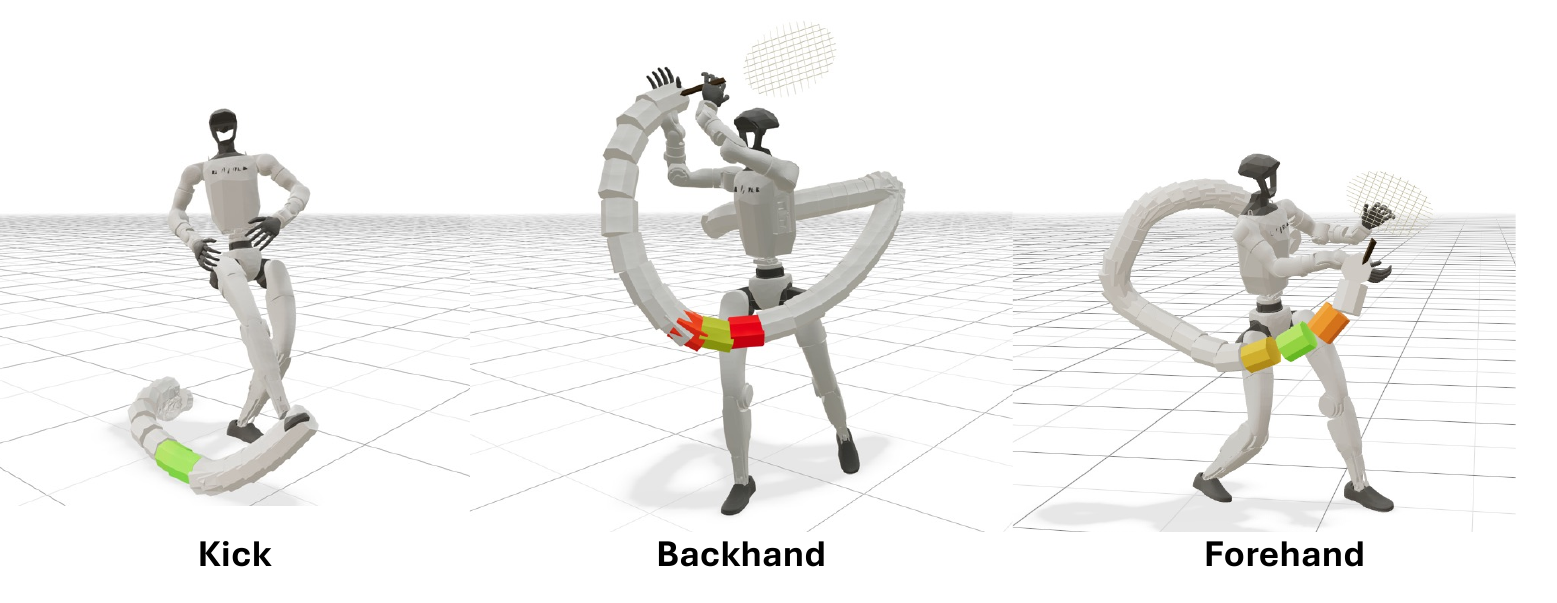}
  \caption{
  \textbf{Training Rewards.} \method learns a policy given a nominal target point and motion reference from a human demonstration (\cref{sec:motions}). Motions and contact points are randomly sampled throughout training (\cref{sec:commands}); position, velocity, and orientation rewards are assigned during the duration of contact. Segments in each motion tubule are color-coded by reward value.}
  \label{fig:motion_tubules}
\end{figure}

Equipped with our action library $\mathbb{A}$, we next learn a policy that computes a trajectory as a function of a chosen goal (\cref{fig:learning_architecture}) and is robust to novel target locations and trajectories at test-time (\cref{fig:motion_tubules}) .

\paragraph{Training}
\label{sec:dynamic_retargeting}
We use Proximal Policy Optimization \cite{schulman2017proximal} to train our policy in the MJlab simulator. Please refer to Appendix \ref{app:training} for more details on our formulation and hyperparameters. 

\paragraph{Observations}
\label{sec:observations}
Our policy is conditioned on both proprioceptive and chosen goal. The proprioceptive inputs include a history of the robot's joint positions, joint velocities, angular velocity, projected gravity vector, and previous action, whose history length we set to 1 sample. 
In addition, the policy receives the chosen goal; target positions and orientations are passed in the robot's local (anchor) frame and target velocities are passed in the world frame, along with the reference trajectory of the chosen action $A_i(t) = \{q_{t}\}_i$. Finally, the critic receives additional privileged observations: robot center of mass position and velocity as well as each of the robot links' positions and orientations, which are detailed in Appendix \ref{app:observations}. 

\paragraph{Actions}
\label{sec:actions}
Our actor outputs position setpoints for each joint to be tracked by PD controllers as introduced by \cite{rudin2022learning} and explored further by \cite{bronars2026tune}. 
Details, including gains, are given in \cref{app:actions}. 

\paragraph{Rewards.}
\label{sec:rewards}

The robot needs to learn to accomplish the goal, and we regularize the solution by imposing that it also imitates the reference motion. Thus, the reward has two main components: trajectory tracking and goal achievement. The tracking component of the reward is taken from \cite{peng2018deepmimic} including positions, orientations, and velocities of robot key points, penalties for self collisions and reaching joint limits, and regularization on action rate. For position and velocity achievement we use 
\begin{equation}
    r_{p, \nu} = w_{\{p, \nu\}}\exp(-\frac{||\{p, \nu\}^* - \{p, \nu\}||_2^2}{\sigma_{\{p, \nu\}}^2})\mathbbm{1}_{\tau \in \Omega}
\end{equation}
and for orientation,
\begin{equation}
    r_{n} = w_{n}\exp(-\frac{1-\langle n^*, n\rangle}{\sigma_{n}^2})\mathbbm{1}_{\tau \in \Omega}.
\end{equation}
where $\langle \cdot, \cdot \rangle$ indicates the standard Euclidean inner product. By decreasing $\sigma_{\{p, \nu, n\}}$ we get a tighter position reward around the goal position if we want stricter adherence to the goal requirement. This allows us to match specific orientation axes, and generalizes to exact orientation matching by using two of these constraints to uniquely define an orientation in $SO(3)$. Notably, our approach does not require any additional reward tuning, besides optionally scaling achievement rewards according to the size of the contact window $\Omega$; see Appendix \ref{app:ablations} for more details.

\paragraph{Action Goals.}
\label{sec:commands}

Our framework accommodates weighted sampling to prioritize motions that are more complex and thus more difficult to learn --- we currently sample all motion classes equally. Action selection relies on assigning heuristic difficulty weights to each motion, which proportionally dictate the exact distribution of motions in a finite pool. Sampling from this pool without replacement during training yields a deterministic motion count, thereby stabilizing the training process.
Conditioned on the motion choice \(i\), we sample $p_i \sim \mathcal{N}(p^*_i,\Sigma),$ and sample \(\nu_i\) and \(n_i\) analogously. Each motion primitive induces a confidence region
\begin{align}
    C_i(\delta)=\left\{
p \in \mathbb{R}^3 :
(p-p^*_i)^\top \Sigma^{-1}(p-p^*_i)
\leq r_\delta^2
\right\},
\end{align}
where \(r_\delta^2\) is chosen such that
\begin{align}
    \mathbb{P}_{p\sim \mathcal{N}(p^*_i,\Sigma)}(p\in C_i(\delta)) = 1-\delta.
\end{align}
where $\delta=0.26$. We implicitly ablate this parameter in Appendix \ref{app:ablations} when we ablate $\sigma_{\{x,y,z\}}$ and find our method is robust to even large variations in the coverage parameter. We choose the motion primitives so that these regions cover the workspace, $W \subseteq \bigcup_{i=1}^K C_i(\delta),$ as illustrated in \cref{fig:pose_coverage}. Orientations and velocities are sampled similarly with their own desired workspaces. 

\paragraph{Curriculum}
 We use reference state initialization similar to \cite{liao2025beyondmimic} to preferentially select more difficult motion phases to initialize to during training. However, when a motion is successfully completed, we do not immediately restart the environment, but instead pause the action at the final frame for a randomized period, then sample a new motion and goal for the robot to perform. This gives the robot an implicit curriculum; at the beginning of training when the robot is initially learning to imitate the motion, adaptive sampling allows us to focus on training the motion itself. However, once the motion is mastered to a degree that the robot no longer loses its balance, we allow the previous motion to act as the initialization for the next one. This way,  training mirrors the situation in the field, allowing the policy to learn good recovery behaviors and transitions between motions. 


\begin{algorithm}[t]
\small
\caption{Action Selection and Goal/Timing Estimation (\cref{sec:motion-selection-and-planning}).}
\label{alg:action_selection}
\begin{algorithmic}[1]
\State \((\hat{x}_b,\hat{v}_b) \gets \mathrm{KalmanEstimate}()\), \quad
       \(\hat{x}_b(t) \gets F(\hat{x}_b,\hat{v}_b)\)

\If{\(\neg A_{\mathrm{locked}}\)}
    \ForAll{\(a_i \in \mathbb{A}\)}
        \State \(\hat{t}_i \gets \arg\min_{t\in[0,T]}\|\hat{x}_b(t)-p_i^*\|_2^2\), \quad
               \(\hat{p}_i \gets \hat{x}_b(\hat{t}_i)\)
    \EndFor
    \State \(i^* \gets \arg\min_i \|\hat{p}_i-p_i^*\|_2\)
    \State \((a^*,\hat{p},\hat{t}) \gets (a_{i^*},\hat{p}_{i^*},\hat{t}_{i^*})\)
    \If{\(\hat{t}<t^*_{i^*}\)}
        \State \(A_{\mathrm{locked}}\gets\mathrm{true}\), \quad \(A_{\mathrm{lock}}\gets a^*\)
    \EndIf
\Else
    \State \(a^*\gets A_{\mathrm{lock}}\), and let \(i^*\) be its index
    \State \(\hat{t} \gets \arg\min_{t\in[0,T]}\|\hat{x}_b(t)-p_{i^*}^*\|_2^2\), \quad
           \(\hat{p}\gets \hat{x}_b(\hat{t})\)
\EndIf

\State \Return \(a^*,\hat{p},\hat{t}\)
\end{algorithmic}
\end{algorithm}

\section{Motion Selection and Planning}
\label{sec:estimation}
\label{sec:motion-selection-and-planning}
The policy from \cref{sec:policy_learning} executes a motion once it is handed a
goal $G = (p,\nu,n,t)$. The job of the planning stack is simple: at every
control step, decide \emph{which} action to play and supply the goal the policy
expects (\cref{fig:method}). This breaks into three steps --- measure the state of
the object being interacted with (e.g. the ball), predict where it will go, and select the action that best deals with that predicted trajectory.

\paragraph{State estimation and prediction.}
We track the dynamic object of interaction (ball, box, etc.) with an OptiTrack
motion-capture system. From the noisy position measurements, a Kalman filter
\cite{welch1995introduction} estimates the object's position and velocity
$(\hat{x}_b,\hat{v}_b)$. We roll this state forward with a hybrid rigid-body model
\cite{nguyen2025high} that combines ballistic flight with bounces:
\begin{align}
F: \begin{cases}
a_b = g, & x_b \notin S \quad\text{(free flight)},\\
v_b^+ = C\,v_b^-, & x_b \in S \quad\text{(bounce)},
\end{cases}
\end{align}
where $g$ is gravitational acceleration, $C$ is an estimated coefficient of
restitution, and $S=\{x_b : x_{b,z}=0\}$ is the ground plane. Propagating the
estimated state through $F$ yields a predicted trajectory $\hat{x}_b(t)$ over the
next $N$ time steps.

\paragraph{Action selection.}
Each candidate action $A_i \in \mathbb{A}$ carries a nominal contact point $p^*_i$
and contact time $t^*_i$, the lead time from motion start to impact
(\cref{sec:motions}). We select the action whose nominal contact point best matches
the predicted trajectory:
\begin{equation}
i^* = \arg\min_{i} \; \min_{t\in[0,T]} \; \|\hat{x}_b(t)-p^*_i\|_2^2 .
\end{equation}
The inner minimization finds, for each action, the time $\hat{t}_i$ at which the ball
passes closest to $p^*_i$; the outer minimization picks the action $i^*$ whose closest
approach is smallest. Note that if $\hat{t}_i < t^{*}_i$, then no action is executed. The target point sent to the policy is the predicted ball
position at that time, $\hat{p}=\hat{x}_b(\hat{t}_{i^*})$, while the velocity direction $\nu$ and orientation $n$ are chosen heuristically. The spatial target is refreshed every step as new measurements arrive. Algorithm
\ref{alg:action_selection} gives the full procedure.

\paragraph{Triggering.}
Because the spatial target keeps updating, the action choice could oscillate as the
prediction is refined. To prevent this, once the predicted time-to-contact drops below
the chosen motion's lead time $t^*_{i^*}$, we lock the action choice and initiate the
motion; the spatial target continues to update until contact.

\section{Experimental Results}
\label{sec:result}

\begin{figure}[]
  \centering
  \includegraphics[width=\linewidth]{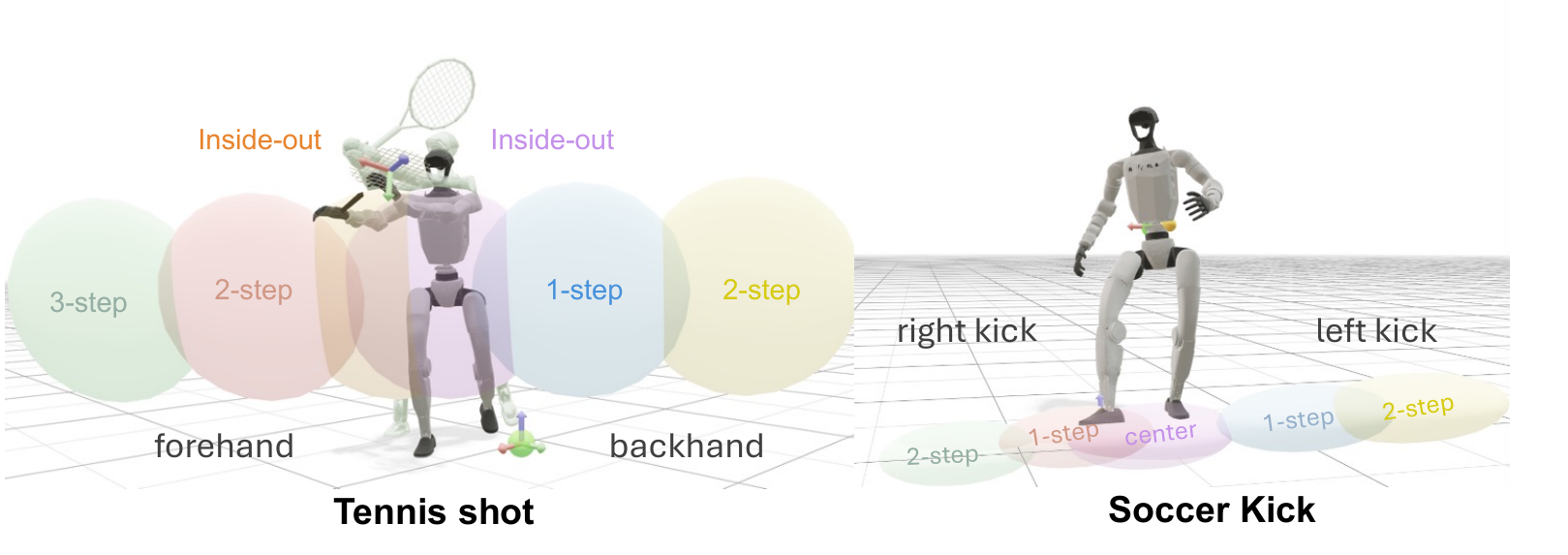}
  \caption{
  \textbf{\method Space Coverage.}
  Our motion abstraction formulation allows us to cover a wide task space. Each colored sphere corresponds to a distinct reference human motion demonstration, and is centered around the point of contact. During training (\cref{sec:policy_learning}) we randomly sample points around each point of contact to provide the learning algorithm with a diverse set of possible ball trajectories. The radius of each sphere represents the standard deviation of the sampling. Notice that the primary actions here are a Cartesian product of the type of shot (e.g. forehand and backhand) and the number of steps needed to reach the proper position to execute the shot.
  }
  \label{fig:pose_coverage}
\end{figure}

We implemented our method on a G1 humanoid robot and carried out experiments to assess whether our learning paradigm is successful in practice. Concretely, we wish to assess three properties of \method: that we can cover a large task space from a few locally varied actions, that our task abstraction gives us robustness in deployment to novel ball trajectories and novel targets, 
and that the combination of these two ideas leads to successful hardware deployment without requiring extensive reward tuning or hyperparameter selection.

\paragraph{Task Space Coverage.}
\method abstracts dynamic motions in a way that enables us to cover a large task space with only a few actions, which we show in ~\cref{fig:pose_coverage}. We define the task space as a 3D volume centered around the demonstration's contact point, as opposed to previous approaches~\cite{su2025hitter} in which the task space is constrained to be a heuristically defined plane. In \cref{tab:sota_comparison} we compare \method to current state-of-the-art humanoid policies in simulation. The generalized success rate (GSR) over this volume is 93\% for ballistic hitting (tennis) and 98\% for box pick-and-place, using sizes equal to those selected for comparison in \cite{wang2026humanx} and using identical metrics for success rate, generalized success rate, and target position error. Pairing variation in 3D contact location with the correct demonstration action yields a policy robust to changes in the 3D target. Notably, from less than 10 demonstrations we are able to achieve successful actions over the same 3D motion volume, without extensive data augmentations or reward tuning as in current methods.

\begin{figure}[]
  \centering
  \includegraphics[width=\linewidth]{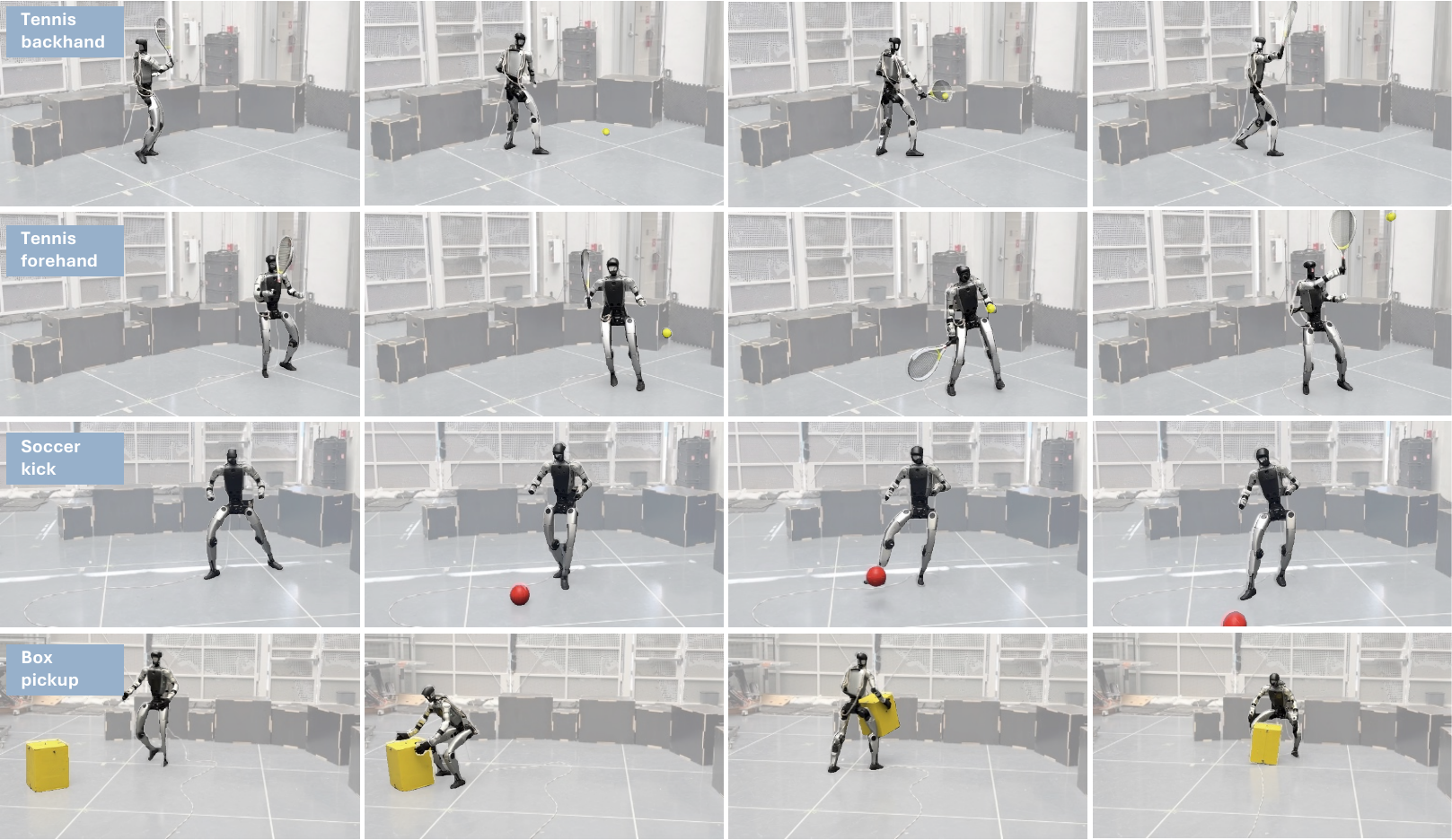}
  \caption{
  \textbf{Qualitative Results.} 
  We train \method on a small set of reference demonstrations of tennis shots, soccer kicks, and box pick-and-place. By randomizing incoming target trajectories and positions, we are able to generalize to unseen target locations in deployment. 
  We show successful execution of tennis shots (1st and 2nd row), soccer kicks (3rd row), and box pick-and-place (4th row). The columns show samples of the robot's pose during task execution.}
  \label{fig:qualitative_results}
\end{figure}

\paragraph{Hardware results.}
\label{hardware}

We report hardware results of our policy in \cref{tab:hardware_results_main}. We deploy our controller on a 27-dof G1 humanoid robot, running the policy onboard, and feed OptiTrack data through ethernet. We use a motion capture system detailed in Appendix \ref{app:estimation} to estimate ball positions, get forward trajectory projections, and feed the target positions to the policy at 50hz. Figure \ref{fig:qualitative_results} and the accompanying video showcase the robot returning balls thrown and kicked with speeds of up to 8m/s and distances up to 2m laterally from the robot. We carried out experiments involving different speed targets. For slow moving targets we have a perfect 100\% hit rate which deteriorates slowly as the speed of the incoming balls increases. We measure 40-70\% at ball toss speeds of 4-8 m/s. Because our task abstraction $G^* = (p^*,\nu^*,n^*, t^*)$ is agnostic to speed (which is handled by the higher-level action selection), all of these scenarios are `in distribution', so the robot fails gracefully, remains standing, and continues to be able to return shots. 

\begin{table}[h]
\caption{\textbf{Hardware Results.} Success rates over 20 trials per condition.}
\label{tab:hardware_results_main}
\centering
\setlength{\tabcolsep}{3.5pt}
\begin{tabularx}{\textwidth}{l CCC @{}}
\toprule
 & \makecell{Tennis SR}
 & \makecell{Soccer SR}
 & \makecell{Box Pick-and-Place SR} \\
\midrule
\textbf{Slow (0-4 m/s)}
    & 1.00
    & 1.00
    & 0.60 \\
\textbf{Fast (4-8 m/s)}
    & 0.45
    & 0.70
    & n/a \\
\bottomrule
\end{tabularx}
\end{table}

\begin{table}[]
\centering
\caption{\textbf{Simulation Result Comparison.}
We compare \method against the state-of-the art. We report success rate (SR); the rate at which the projectile is hit by the robot or the box is picked to a height of within 10cm of the target (waist) position with nominal ball trajectories/box positions from the original demonstration; generalized success rate (GSR); where the trajectory initial conditions/box positions are randomized either uniformly in a ball of size 0.3m in the ballistic case, or uniformly with a distance of up to 3m in the semicircular area in front of the robot in the box case; and target position error ($e_b$); the distance between the correct and actual interaction point (contact with box or ball). XMimic has the highest SR for ballistic hitting, but \method has a higher GSR, suggesting that it is better able to generalize to novel trajectories. \method performs best across all other metrics.}
\label{tab:sota_comparison}

\small
\setlength{\tabcolsep}{3.5pt}
\renewcommand{\arraystretch}{1.05}

\begin{tabular*}{\textwidth}{@{\extracolsep{\fill}} l c c c c c c @{}}
\toprule
\addlinespace[0.15em]
\textbf{Method}
& \multicolumn{3}{c}{\textbf{Ballistic Hitting}}
& \multicolumn{3}{c}{\textbf{Box Pick-and Place}} \\
\cmidrule(lr){2-4}
\cmidrule(lr){5-7}

& SR$\uparrow$
& GSR$\uparrow$
& $e_b$ (m)$\downarrow$
& SR$\uparrow$
& GSR$\uparrow$
& $e_b$ (cm)$\downarrow$ \\

\addlinespace[0.15em]
\midrule
\addlinespace[0.15em]

SkillMimic~\cite{wang2025skillmimic}
& 68.2\% & 30.9\% & 0.48 {\tiny (0.13)}
& 0.4\% & 0.0\% & 28.3 {\tiny (2.13)} \\

OmniRetarget~\cite{yang2025omniretarget}
& 75.8\% & 20.4\% & 0.15 {\tiny (0.10)}
& 0.0\% & 0.0\% & 13.3 {\tiny (1.57)} \\

HDMI~\cite{weng2025hdmi}
& 90.0\% & 25.3\% & 0.07 {\tiny (0.02)}
& 95.8\% & 1.8\% & 13.1 {\tiny (1.28)} \\

HumanX~\cite{wang2026humanx}
& \textbf{100\%} & 90.6\% & 0.09 {\tiny (0.06)}
& 99.3\% & 96.3\% & 8.67 {\tiny (0.91)} \\

\addlinespace[0.15em]
\midrule
\addlinespace[0.15em]

TaskNPoint
& 99.5\% & \textbf{93.0\%} & \textbf{0.02 {\tiny (0.01)}}
& \textbf{100.0\%} & \textbf{98.0\%} & \textbf{4.85 {\tiny (3.65)}} \\

\addlinespace[0.15em]
\bottomrule
\end{tabular*}
\end{table}

\begin{figure}[h]
  \centering
  \includegraphics[width=\linewidth]{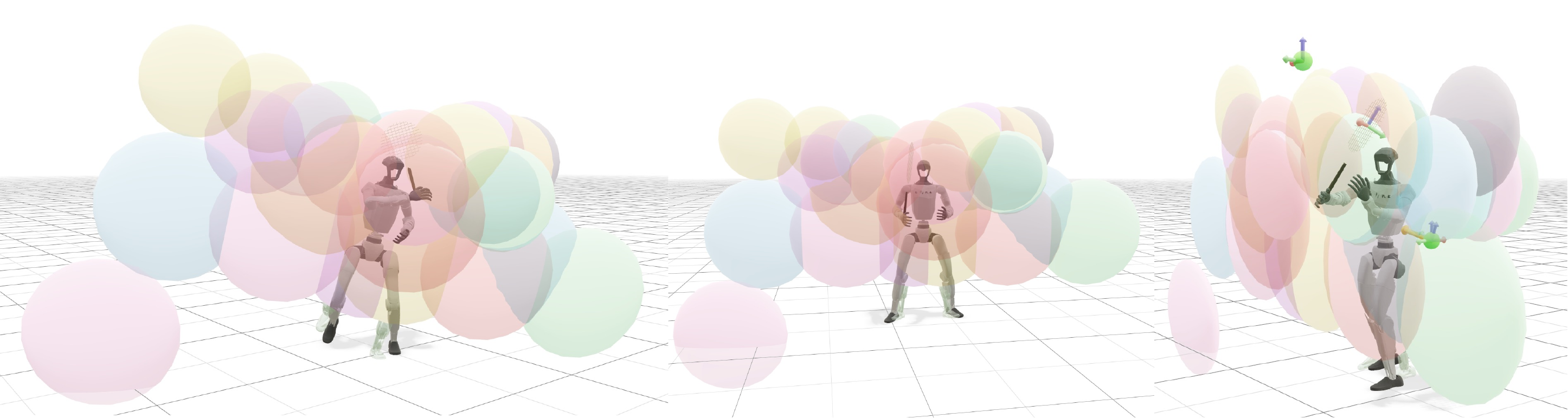}
  \caption{
  \textbf{MLE poses}. We test how well \method scales with the number of demonstrations by training on MLE-estimated human poses from in-the-wild multi-view data of tennis practices and matches (\cref{sec:motion_recon}). Here we visualize the nominal point sampling volume during training. Each colored sphere is centered around the nominal point and corresponds to a distinct human demonstration, and the radius of each sphere corresponds to one standard deviation of the sampling area during training. We note that by increasing the number of demonstrations we are able to cover more shots, including volleys and overheads, and that many of the demonstration volumes overlap with each other, resulting in the ability to perform different actions on the same 3D point in space.
  }
  \label{fig:mle_vis_pose}
\end{figure}

\paragraph{Generalizing to Novel Tasks: Box Pick-and-Place.}
How general is the abstraction of motion as a 3D conditioning point and contact time? We test this with a box pickup task, wherein we place boxes randomly in a 3m radius in front of the robot and task the robot to pick up the box and return it to its starting location, which we show in the accompanying video as well. We further discuss hardware details, limitations, and performance in Appendix \ref{app:hardware_results}.

\paragraph{Ablations.}

We run several ablations on our parameter selection and see that our policy is not only performant in the nominal setup, but robust to considerable variation in parameters from phase density, $\tau(t^{*,+})-\tau(t^{*,-})$, to target nominal distance $||p_i-p^*_i||$. We see graceful degradation of quality outside our best parameter choices, where failure rate remains consistently less than 10\% and the average target position error falls until the robot is just performing its nominal motion. See our discussion in Appendix \ref{app:ablations}.

\paragraph{Training On In-the-Wild Data.}

We test the ability of \method to handle additional motions by retraining the policy on up to 34 discrete motions (shown in \cref{fig:mle_vis_pose}) and report our results in \cref{app:motion_count}. Overall success rate on 34 motions is 91\%, with an average target position error of 4.9cm. The phase error remains at 0.5, unchanged from the baseline. This suggests that although we demonstrate successful deployment of a limited number of motions, our method scales well to more complex motion libraries, and does not sacrifice its quality. 

\section{Conclusion}
\label{sec:conclusion}

We introduced TaskNPoint, an abstraction methodology and training pipeline for teaching dynamic skills to humanoid robots. We observe that the space of human motions is intrinsically discrete, composed of a suite of skills, each of which is dynamically adapted to environmental conditions and continuously varied in order to achieve a given goal. Through an intentionally simple implementation of our idea, we find that we can teach dynamic loco-manipulation skills such as tennis and soccer play and box picking and placing with only seconds of training data from a coach, and under an hour of training per task, all on a single GPU. We believe this direction offers a scalable and robust paradigm for humanoid learning of dynamic actions such as those found in sports. We expect future work in this direction to approach the problem of environmental conditioning for motion and target selection, along with targeted modeling of key interactions.
\section{Limitations}
\label{sec:limitations}

\method's loco-manipulation abilities are hindered by the lack of force feedback in training: we assume that the tasks we perform do not require this feedback for success. We believe that a direction more similar to \cite{wang2026humanx} could be better here, in which we train with the simulated object-to-be-manipulated and therefore allow the policy to learn recovery behaviors, tactile feedback, and sensor robustness. However, this comes at a cost of complex data generation, longer training, and assumptions about the types of objects with which we will be interacting. Additionally, \method assumes ground-truth estimator accuracy, and in practice the majority of our hardware deployment failure cases are due to imprecise estimates of ball trajectories. The relaxation of our hierarchical abstraction could provide a promising way forward at cost of generality, for instance, by modeling the estimator in the training pipeline. 



\clearpage
\acknowledgments{Thanks to Lizhi Yang, Sergio Esteban, Nima Rahmanian, and Georgia Gkioxari for their advice. Thanks to Damiano Marsili for reviewing. This work is supported by the Technology Innovation Institute (TII) and the National Science Foundation Graduate Research Fellowship Program under Grant No. 2139433. Data of the Caltech Tennis Team was obtained through explicit IRB approved consent.}


\bibliography{tasknpoint}  

@article{su2025hitter,
  title={Hitter: A humanoid table tennis robot via hierarchical planning and learning},
  author={Su, Zhi and Zhang, Bike and Rahmanian, Nima and Gao, Yuman and Liao, Qiayuan and Regan, Caitlin and Sreenath, Koushil and Sastry, S Shankar},
  journal={arXiv preprint arXiv:2508.21043},
  year={2025}
}

@article{allshire2025visual,
  title={Visual imitation enables contextual humanoid control},
  author={Allshire, Arthur and Choi, Hongsuk and Zhang, Junyi and McAllister, David and Zhang, Anthony and Kim, Chung Min and Darrell, Trevor and Abbeel, Pieter and Malik, Jitendra and Kanazawa, Angjoo},
  journal={arXiv preprint arXiv:2505.03729},
  year={2025}
}

@article{zhang2026learning,
  title={Learning athletic humanoid tennis skills from imperfect human motion data},
  author={Zhang, Zhikai and Lu, Haofei and Lian, Yunrui and Chen, Ziqing and Liu, Yun and Lin, Chenghuai and Xue, Han and Zeng, Zicheng and Qi, Zekun and Zheng, Shaolin and others},
  journal={arXiv preprint arXiv:2603.12686},
  year={2026}
}

@article{peng2018deepmimic,
  title={Deepmimic: Example-guided deep reinforcement learning of physics-based character skills},
  author={Peng, Xue Bin and Abbeel, Pieter and Levine, Sergey and Van de Panne, Michiel},
  journal={ACM Transactions On Graphics (TOG)},
  volume={37},
  number={4},
  pages={1--14},
  year={2018},
  publisher={ACM New York, NY, USA}
}

@article{liao2025beyondmimic,
  title={Beyondmimic: From motion tracking to versatile humanoid control via guided diffusion},
  author={Liao, Qiayuan and Truong, Takara E and Huang, Xiaoyu and Gao, Yuman and Tevet, Guy and Sreenath, Koushil and Liu, C Karen},
  journal={arXiv preprint arXiv:2508.08241},
  year={2025}
}

@article{wang2026humanx,
  title={HumanX: Toward Agile and Generalizable Humanoid Interaction Skills from Human Videos},
  author={Wang, Yinhuai and Zhao, Qihan and Lau, Yuen Fui and Yu, Runyi and Tsui, Hok Wai and Chen, Qifeng and Wang, Jingbo and Pang, Jiangmiao and Tan, Ping},
  journal={arXiv preprint arXiv:2602.02473},
  year={2026}
}

@article{araujo2025retargeting,
  title={Retargeting matters: General motion retargeting for humanoid motion tracking},
  author={Araujo, Joao Pedro and Ze, Yanjie and Xu, Pei and Wu, Jiajun and Liu, C Karen},
  journal={arXiv preprint arXiv:2510.02252},
  year={2025}
}

@misc{xu2022vitposesimplevisiontransformer,
      title={ViTPose: Simple Vision Transformer Baselines for Human Pose Estimation}, 
      author={Yufei Xu and Jing Zhang and Qiming Zhang and Dacheng Tao},
      year={2022},
      eprint={2204.12484},
      archivePrefix={arXiv},
      primaryClass={cs.CV},
      url={https://arxiv.org/abs/2204.12484}, 
}

@misc{ravi2024sam2segmentimages,
      title={SAM 2: Segment Anything in Images and Videos}, 
      author={Nikhila Ravi and Valentin Gabeur and Yuan-Ting Hu and Ronghang Hu and Chaitanya Ryali and Tengyu Ma and Haitham Khedr and Roman Rädle and Chloe Rolland and Laura Gustafson and Eric Mintun and Junting Pan and Kalyan Vasudev Alwala and Nicolas Carion and Chao-Yuan Wu and Ross Girshick and Piotr Dollár and Christoph Feichtenhofer},
      year={2024},
      eprint={2408.00714},
      archivePrefix={arXiv},
      primaryClass={cs.CV},
      url={https://arxiv.org/abs/2408.00714}, 
}

@misc{pavlakos2019smplx,
      title={Expressive Body Capture: 3D Hands, Face, and Body from a Single Image}, 
      author={Georgios Pavlakos and Vasileios Choutas and Nima Ghorbani and Timo Bolkart and Ahmed A. A. Osman and Dimitrios Tzionas and Michael J. Black},
      year={2019},
      eprint={1904.05866},
      archivePrefix={arXiv},
      primaryClass={cs.CV},
      url={https://arxiv.org/abs/1904.05866}, 
}

@misc{wang2025prompthmrpromptablehumanmesh,
      title={PromptHMR: Promptable Human Mesh Recovery}, 
      author={Yufu Wang and Yu Sun and Priyanka Patel and Kostas Daniilidis and Michael J. Black and Muhammed Kocabas},
      year={2025},
      eprint={2504.06397},
      archivePrefix={arXiv},
      primaryClass={cs.CV},
      url={https://arxiv.org/abs/2504.06397}, 
}

@misc{teed2022droidslamdeepvisualslam,
      title={DROID-SLAM: Deep Visual SLAM for Monocular, Stereo, and RGB-D Cameras}, 
      author={Zachary Teed and Jia Deng},
      year={2022},
      eprint={2108.10869},
      archivePrefix={arXiv},
      primaryClass={cs.CV},
      url={https://arxiv.org/abs/2108.10869}, 
}

@misc{bhat2023zoedepthzeroshottransfercombining,
      title={ZoeDepth: Zero-shot Transfer by Combining Relative and Metric Depth}, 
      author={Shariq Farooq Bhat and Reiner Birkl and Diana Wofk and Peter Wonka and Matthias Müller},
      year={2023},
      eprint={2302.12288},
      archivePrefix={arXiv},
      primaryClass={cs.CV},
      url={https://arxiv.org/abs/2302.12288}, 
}

@article{ze2025twist2,
  title={Twist2: Scalable, portable, and holistic humanoid data collection system},
  author={Ze, Yanjie and Zhao, Siheng and Wang, Weizhuo and Kanazawa, Angjoo and Duan, Rocky and Abbeel, Pieter and Shi, Guanya and Wu, Jiajun and Liu, C Karen},
  journal={arXiv preprint arXiv:2511.02832},
  year={2025}
}

@article{ma2025learning,
  title={Learning coordinated badminton skills for legged manipulators},
  author={Ma, Yuntao and Cramariuc, Andrei and Farshidian, Farbod and Hutter, Marco},
  journal={Science robotics},
  volume={10},
  number={102},
  pages={eadu3922},
  year={2025},
  publisher={American Association for the Advancement of Science}
}

@article{schulman2017proximal,
  title={Proximal policy optimization algorithms},
  author={Schulman, John and Wolski, Filip and Dhariwal, Prafulla and Radford, Alec and Klimov, Oleg},
  journal={arXiv preprint arXiv:1707.06347},
  year={2017}
}

@inproceedings{rudin2022learning,
  title={Learning to walk in minutes using massively parallel deep reinforcement learning},
  author={Rudin, Nikita and Hoeller, David and Reist, Philipp and Hutter, Marco},
  booktitle={Conference on robot learning},
  pages={91--100},
  year={2022},
  organization={PMLR}
}

@article{zakka2026mjlab,
  title={mjlab: A Lightweight Framework for GPU-Accelerated Robot Learning},
  author={Zakka, Kevin and Liao, Qiayuan and Yi, Brent and Lay, Louis Le and Sreenath, Koushil and Abbeel, Pieter},
  journal={arXiv preprint arXiv:2601.22074},
  year={2026}
}

@article{weng2025hdmi,
  title={Hdmi: Learning interactive humanoid whole-body control from human videos},
  author={Weng, Haoyang and Li, Yitang and Sobanbabu, Nikhil and Wang, Zihan and Luo, Zhengyi and He, Tairan and Ramanan, Deva and Shi, Guanya},
  journal={arXiv preprint arXiv:2509.16757},
  year={2025}
}

@article{yang2025omniretarget,
  title={Omniretarget: Interaction-preserving data generation for humanoid whole-body loco-manipulation and scene interaction},
  author={Yang, Lujie and Huang, Xiaoyu and Wu, Zhen and Kanazawa, Angjoo and Abbeel, Pieter and Sferrazza, Carmelo and Liu, C Karen and Duan, Rocky and Shi, Guanya},
  journal={arXiv preprint arXiv:2509.26633},
  year={2025}
}

@inproceedings{wang2025skillmimic,
  title={Skillmimic: Learning basketball interaction skills from demonstrations},
  author={Wang, Yinhuai and Zhao, Qihan and Yu, Runyi and Tsui, Hok Wai and Zeng, Ailing and Lin, Jing and Luo, Zhengyi and Yu, Jiwen and Li, Xiu and Chen, Qifeng and others},
  booktitle={Proceedings of the IEEE/CVF Conference on Computer Vision and Pattern Recognition},
  pages={17540--17549},
  year={2025}
}

@article{calinon2007teacher,
  author    = {Calinon, Sylvain and Billard, Aude G.},
  title     = {What is the Teacher's Role in Robot Programming by Demonstration? -- Toward Benchmarks for Improved Learning},
  journal   = {Interaction Studies},
  volume    = {8},
  number    = {3},
  pages     = {441--464},
  year      = {2007},
  doi       = {10.1075/is.8.3.08cal},
  publisher = {John Benjamins}
}

@incollection{nehaniv2002correspondence,
  author    = {Nehaniv, Chrystopher L. and Dautenhahn, Kerstin},
  title     = {The Correspondence Problem},
  booktitle = {Imitation in Animals and Artifacts},
  editor    = {Dautenhahn, Kerstin and Nehaniv, Chrystopher L.},
  pages     = {41--61},
  publisher = {MIT Press},
  address   = {Cambridge, MA},
  year      = {2002},
  isbn      = {9780262042031}
}

@article{haarnoja2024learning,
  title={Learning agile soccer skills for a bipedal robot with deep reinforcement learning},
  author={Haarnoja, Tuomas and Moran, Ben and Lever, Guy and Huang, Sandy H and Tirumala, Dhruva and Humplik, Jan and Wulfmeier, Markus and Tunyasuvunakool, Saran and Siegel, Noah Y and Hafner, Roland and others},
  journal={Science Robotics},
  volume={9},
  number={89},
  pages={eadi8022},
  year={2024},
  publisher={American Association for the Advancement of Science}
}

@article{wu2026perceptive,
  title={Perceptive humanoid parkour: Chaining dynamic human skills via motion matching},
  author={Wu, Zhen and Huang, Xiaoyu and Yang, Lujie and Zhang, Yuanhang and Sreenath, Koushil and Chen, Xi and Abbeel, Pieter and Duan, Rocky and Kanazawa, Angjoo and Sferrazza, Carmelo and others},
  journal={arXiv preprint arXiv:2602.15827},
  year={2026}
}

@article{wang2025beamdojo,
  title={Beamdojo: Learning agile humanoid locomotion on sparse footholds},
  author={Wang, Huayi and Wang, Zirui and Ren, Junli and Ben, Qingwei and Huang, Tao and Zhang, Weinan and Pang, Jiangmiao},
  journal={arXiv preprint arXiv:2502.10363},
  year={2025}
}

@article{hwangbo2018per,
  title={Per-contact iteration method for solving contact dynamics},
  author={Hwangbo, Jemin and Lee, Joonho and Hutter, Marco},
  journal={IEEE Robotics and Automation Letters},
  volume={3},
  number={2},
  pages={895--902},
  year={2018},
  publisher={IEEE}
}

@article{makoviychuk2021isaac,
  title={Isaac gym: High performance gpu-based physics simulation for robot learning},
  author={Makoviychuk, Viktor and Wawrzyniak, Lukasz and Guo, Yunrong and Lu, Michelle and Storey, Kier and Macklin, Miles and Hoeller, David and Rudin, Nikita and Allshire, Arthur and Handa, Ankur and others},
  journal={arXiv preprint arXiv:2108.10470},
  year={2021}
}

@article{yu2021human,
  title={Human dynamics from monocular video with dynamic camera movements},
  author={Yu, Ri and Park, Hwangpil and Lee, Jehee},
  journal={ACM Transactions on Graphics (TOG)},
  volume={40},
  number={6},
  pages={1--14},
  year={2021},
  publisher={ACM New York, NY, USA}
}

@inproceedings{luo2024universal,
  title={Universal humanoid motion representations for physics-based control},
  author={Luo, Zhengyi and Cao, Jinkun and Merel, Josh and Winkler, Alexander and Huang, Jing and Kitani, Kris and Xu, Weipeng},
  booktitle={International Conference on Learning Representations},
  volume={2024},
  pages={56766--56782},
  year={2024}
}

@inproceedings{peng2021neural,
  title={Neural body: Implicit neural representations with structured latent codes for novel view synthesis of dynamic humans},
  author={Peng, Sida and Zhang, Yuanqing and Xu, Yinghao and Wang, Qianqian and Shuai, Qing and Bao, Hujun and Zhou, Xiaowei},
  booktitle={Proceedings of the IEEE/CVF conference on computer vision and pattern recognition},
  pages={9054--9063},
  year={2021}
}

@inproceedings{moon2024expressive,
  title={Expressive whole-body 3d gaussian avatar},
  author={Moon, Gyeongsik and Shiratori, Takaaki and Saito, Shunsuke},
  booktitle={European Conference on Computer Vision},
  pages={19--35},
  year={2024},
  organization={Springer}
}

@inproceedings{rajasegaran2022tracking,
  title={Tracking people by predicting 3d appearance, location and pose},
  author={Rajasegaran, Jathushan and Pavlakos, Georgios and Kanazawa, Angjoo and Malik, Jitendra},
  booktitle={Proceedings of the IEEE/CVF conference on computer vision and pattern recognition},
  pages={2740--2749},
  year={2022}
}

@inproceedings{luvizon20182d,
  title={2d/3d pose estimation and action recognition using multitask deep learning},
  author={Luvizon, Diogo C and Picard, David and Tabia, Hedi},
  booktitle={Proceedings of the IEEE conference on computer vision and pattern recognition},
  pages={5137--5146},
  year={2018}
}

@inproceedings{rajasegaran2023benefits,
  title={On the benefits of 3d pose and tracking for human action recognition},
  author={Rajasegaran, Jathushan and Pavlakos, Georgios and Kanazawa, Angjoo and Feichtenhofer, Christoph and Malik, Jitendra},
  booktitle={Proceedings of the IEEE/CVF conference on computer vision and pattern recognition},
  pages={640--649},
  year={2023}
}

@inproceedings{yuan2021simpoe,
  title={Simpoe: Simulated character control for 3d human pose estimation},
  author={Yuan, Ye and Wei, Shih-En and Simon, Tomas and Kitani, Kris and Saragih, Jason},
  booktitle={Proceedings of the IEEE/CVF conference on computer vision and pattern recognition},
  pages={7159--7169},
  year={2021}
}

@article{yuan2023learning,
  title={Learning physically simulated tennis skills from broadcast videos},
  author={Yuan, Ye and Makoviychuk, Viktor and Guo, Y and Fidler, S and Peng, X and Fatahalian, K},
  journal={ACM Trans. Graph},
  volume={42},
  number={4},
  pages={66},
  year={2023}
}

@inproceedings{ugrinovic2024multiphys,
  title={Multiphys: Multi-person physics-aware 3d motion estimation},
  author={Ugrinovic, Nicolas and Pan, Boxiao and Pavlakos, Georgios and Paschalidou, Despoina and Shen, Bokui and Sanchez-Riera, Jordi and Moreno-Noguer, Francesc and Guibas, Leonidas},
  booktitle={Proceedings of the IEEE/CVF Conference on Computer Vision and Pattern Recognition},
  pages={2331--2340},
  year={2024}
}

@inproceedings{li2022d,
  title={D \&d: Learning human dynamics from dynamic camera},
  author={Li, Jiefeng and Bian, Siyuan and Xu, Chao and Liu, Gang and Yu, Gang and Lu, Cewu},
  booktitle={European Conference on Computer Vision},
  pages={479--496},
  year={2022},
  organization={Springer}
}

@article{chen2026learning,
  title={Learning human-like badminton skills for humanoid robots},
  author={Chen, Yeke and Dong, Shihao and Ji, Xiaoyu and Sun, Jingkai and Luo, Zeren and Zhao, Liu and Zhang, Jiahui and Li, Wanyue and Ma, Ji and Xu, Bowen and others},
  journal={arXiv preprint arXiv:2602.08370},
  year={2026}
}

@article{kim2025physicsfc,
  title={PhysicsFC: Learning User-Controlled Skills for a Physics-Based Football Player Controller},
  author={Kim, Minsu and Jung, Eunho and Lee, Yoonsang},
  journal={ACM Transactions on Graphics (TOG)},
  volume={44},
  number={4},
  pages={1--21},
  year={2025},
  publisher={ACM New York, NY, USA}
}

@article{liu2025humanoid,
  title={Humanoid Whole-Body Badminton via Multi-Stage Reinforcement Learning},
  author={Liu, Chenhao and Jiang, Leyun and Wang, Yibo and Yao, Kairan and Fu, Jinchen and Ren, Xiaoyu},
  journal={arXiv preprint arXiv:2511.11218},
  year={2025}
}

@article{liu2018learning,
  title={Learning basketball dribbling skills using trajectory optimization and deep reinforcement learning},
  author={Liu, Libin and Hodgins, Jessica},
  journal={Acm transactions on graphics (tog)},
  volume={37},
  number={4},
  pages={1--14},
  year={2018},
  publisher={ACM New York, NY, USA}
}

@article{luo2024smplolympics,
  title={Smplolympics: Sports environments for physically simulated humanoids},
  author={Luo, Zhengyi and Wang, Jiashun and Liu, Kangni and Zhang, Haotian and Tessler, Chen and Wang, Jingbo and Yuan, Ye and Cao, Jinkun and Lin, Zihui and Wang, Fengyi and others},
  journal={arXiv preprint arXiv:2407.00187},
  year={2024}
}

@article{ren2025humanoid,
  title={Humanoid Goalkeeper: Learning from Position Conditioned Task-Motion Constraints},
  author={Ren, Junli and Long, Junfeng and Huang, Tao and Wang, Huayi and Wang, Zirui and Jia, Feiyu and Zhang, Wentao and Wang, Jingbo and Luo, Ping and Pang, Jiangmiao},
  journal={arXiv preprint arXiv:2510.18002},
  year={2025}
}

@article{wang2025hierarchical,
  title={A Hierarchical, Model-Based System for High-Performance Humanoid Soccer},
  author={Wang, Quanyou and Zhu, Mingzhang and Hou, Ruochen and Gillespie, Kay and Zhu, Alvin and Wang, Shiqi and Wang, Yicheng and Fernandez, Gaberiel I and Liu, Yeting and Togashi, Colin and others},
  journal={arXiv preprint arXiv:2512.09431},
  year={2025}
}

@inproceedings{zhang2021learning,
  title={Learning motion priors for 4d human body capture in 3d scenes},
  author={Zhang, Siwei and Zhang, Yan and Bogo, Federica and Pollefeys, Marc and Tang, Siyu},
  booktitle={Proceedings of the IEEE/CVF International Conference on Computer Vision},
  pages={11343--11353},
  year={2021}
}

@article{welch1995introduction,
  title={An introduction to the Kalman filter},
  author={Welch, Greg and Bishop, Gary and others},
  year={1995},
  publisher={Chapel Hill, NC, USA}
}

@misc{kanazawa2018hmr,
      title={Beyond Static Features for Temporally Consistent 3D Human Pose and Shape from a Video}, 
      author={Hongsuk Choi and Gyeongsik Moon and Ju Yong Chang and Kyoung Mu Lee},
      year={2021},
      eprint={2011.08627},
      archivePrefix={arXiv},
      primaryClass={cs.CV},
      url={https://arxiv.org/abs/2011.08627}, 
}

@misc{kanazawa2019hmmr,
      title={Learning 3D Human Dynamics from Video}, 
      author={Angjoo Kanazawa and Jason Y. Zhang and Panna Felsen and Jitendra Malik},
      year={2019},
      eprint={1812.01601},
      archivePrefix={arXiv},
      primaryClass={cs.CV},
      url={https://arxiv.org/abs/1812.01601}, 
}

@misc{daniilidis2024tram,
      title={TRAM: Global Trajectory and Motion of 3D Humans from in-the-wild Videos}, 
      author={Yufu Wang and Ziyun Wang and Lingjie Liu and Kostas Daniilidis},
      year={2024},
      eprint={2403.17346},
      archivePrefix={arXiv},
      primaryClass={cs.CV},
      url={https://arxiv.org/abs/2403.17346}, 
}

@misc{yuan2025genmo,
      title={GENMO: A GENeralist Model for Human MOtion}, 
      author={Jiefeng Li and Jinkun Cao and Haotian Zhang and Davis Rempe and Jan Kautz and Umar Iqbal and Ye Yuan},
      year={2025},
      eprint={2505.01425},
      archivePrefix={arXiv},
      primaryClass={cs.GR},
      url={https://arxiv.org/abs/2505.01425}, 
}

@inproceedings{xiaowei2024gvhmr, series={SA ’24},
   title={World-Grounded Human Motion Recovery via Gravity-View Coordinates},
   url={http://dx.doi.org/10.1145/3680528.3687565},
   DOI={10.1145/3680528.3687565},
   booktitle={SIGGRAPH Asia 2024 Conference Papers},
   publisher={ACM},
   author={Shen, Zehong and Pi, Huaijin and Xia, Yan and Cen, Zhi and Peng, Sida and Hu, Zechen and Bao, Hujun and Hu, Ruizhen and Zhou, Xiaowei},
   year={2024},
   month=dec, pages={1–11},
   collection={SA ’24} }

@misc{yeung2025athletepose3d,
      title={AthletePose3D: A Benchmark Dataset for 3D Human Pose Estimation and Kinematic Validation in Athletic Movements}, 
      author={Calvin Yeung and Tomohiro Suzuki and Ryota Tanaka and Zhuoer Yin and Keisuke Fujii},
      year={2025},
      eprint={2503.07499},
      archivePrefix={arXiv},
      primaryClass={cs.CV},
      url={https://arxiv.org/abs/2503.07499}, 
}

@inproceedings{nguyen2025high,
  title={High speed robotic table tennis swinging using lightweight hardware with model predictive control},
  author={Nguyen, David and Cancio, Kendrick D and Kim, Sangbae},
  booktitle={2025 IEEE International Conference on Robotics and Automation (ICRA)},
  pages={15278--15284},
  year={2025},
  organization={IEEE}
}

@article{peng2021amp,
  title={Amp: Adversarial motion priors for stylized physics-based character control},
  author={Peng, Xue Bin and Ma, Ze and Abbeel, Pieter and Levine, Sergey and Kanazawa, Angjoo},
  journal={ACM Transactions on Graphics (ToG)},
  volume={40},
  number={4},
  pages={1--20},
  year={2021},
  publisher={ACM New York, NY, USA}
}

@misc{demler2026caltennis,
      title={CalTennis: Large Multi-View Tennis Video Dataset and Benchmark of Monocular-to-3D Pose Estimation}, 
      author={Ilona Demler and Xinran Xie and Blake Werner and Anna Szczuka and Pietro Perona},
      year={2026},
      eprint={2606.20542},
      archivePrefix={arXiv},
      primaryClass={cs.CV},
      url={https://arxiv.org/abs/2606.20542}, 
}

@article{bronars2026tune,
  title={Tune to learn: How controller gains shape robot policy learning},
  author={Bronars, Antonia and Park, Younghyo and Agrawal, Pulkit},
  journal={arXiv preprint arXiv:2604.02523},
  year={2026}
}

\newpage
\appendix
\section{Notation Table}
\label{app:notation_table}

\begin{table}[h]
\centering
\caption{Notation and Definitions}
\label{tab:notation}
\begin{tabular}{ll}
\toprule
\textbf{Variable} & \textbf{Definition} \\
\midrule
\multicolumn{2}{l}{\textit{Video capture and camera calibration}} \\
\midrule
$N$ & Number of cameras capturing the scene \\
$c^i$ & The $i$-th camera \\
$V^i$ & Video collected by camera $c^i$, containing RGB frames $I_t^i$ \\
$t^i_k$ & Length (total frames or duration) of video $V^i$ \\
$K^i$ & Intrinsic matrix of camera $c^i$ ($\mathbb{R}^{3 \times 4}$) \\
$R^i, T^i$ & Extrinsic rotation ($SO(3)$) and translation ($\mathbb{R}^3$) of camera $c^i$ \\
\midrule
\multicolumn{2}{l}{\textit{Human pose reconstruction}} \\
\midrule
$p$ & Number of people in the scene \\
$M$ & Pre-trained monocular 3D human pose estimation model \\
$H^i$ & SMPL-X pose estimates for $p$ people from video $V^i$ \\
$d^i_t$ & Translation of all $p$ people at time $t$ ($\mathbb{R}^{p \times 3}$) \\
$\phi^i_t$ & Global orientation of all $p$ people at time $t$ ($\mathbb{R}^{p \times 3}$) \\
$\theta^i_t$ & Body pose (joint angles) of all $p$ people at time $t$ ($\mathbb{R}^{p \times 21 \times 3}$) \\
$\beta^i_t$ & Body shape parameters of all $p$ people at time $t$ ($\mathbb{R}^{p \times 10}$) \\
\midrule
\multicolumn{2}{l}{\textit{Multi-view fusion}} \\
\midrule
$J^{(i)}_c, J^{(i)}_w$ & Joint position estimate from camera $i$ in camera / world frame ($\mathbb{R}^3$) \\
$\Sigma^{(i)}_c, \Sigma^{(i)}_w$ & Per-camera joint observation covariance, camera / world frame ($\mathbb{R}^{3 \times 3}$) \\
$P_{\text{MLE}}$ & Maximum-likelihood consensus joint position ($\mathbb{R}^3$) \\
\midrule
\multicolumn{2}{l}{\textit{Reference motion and task}} \\
\midrule
$\mathbb{A} := \{A_1,A_2, \dots\}$ & Set of reference actions $(\mathbb{R}^{n \times T \times K})$ \\ 
$K$ & Number of discrete actions from which the policy selects \\
$T$ & Number of frames in a reference motion \\
$n$ & Number of tracked key points on the robot \\
$\mathbb{G}$ & A TaskNPoint goal defined as the tuple $\{ (p^*,\nu^*,n^*, t^*) \}$ \\
$q^{*}_{1:T}$ & Reference key-point positions ($\mathbb{R}^{n \times T}$) \\
$\tau(t) \in [0,1]$ & Speed-agnostic motion phase, with $\tau(0)=0$ and $\tau(T)=1$ \\
$(t^{*,-}, t^{*,+})$ & Time window of the crucial interaction (e.g., racket–ball contact) \\
$\Omega := (\tau(t^{*,-}), \tau(t^{*,+}))$ & Phase window in which target rewards are active \\
$p^* $ & Target interaction point in the world \\
$\nu^*$ & Target interaction velocity \\
$n^* $ & Target interaction orientation axis \\
\midrule
\multicolumn{2}{l}{\textit{Training and target sampling}} \\
\midrule
$i$ & Action / motion class index \\
$C_i(\delta)$ & Confidence region around a target for motion $i$ \\
$\Sigma_{p, \nu, n} = \mathrm{diag}(\sigma_x, \sigma_y, \sigma_z)_{p, \nu, n}$ & Covariance of target sampling around $p^*, \nu^*, n^*$\\
$\mathcal{W}$ & Covered workspace, $W \subseteq \bigcup_{i=1}^K C_i(\delta),$ \\
$\delta$ & Coverage probability parameter \\
$r_{\{p,\nu\}}, r_n$ & Target position/velocity and orientation reward terms \\
$w_{\{p,\nu\}}, w_n$ & Target reward weights \\
$\sigma_{\{p,\nu\}}, \sigma_n$ & Target reward bandwidths \\

\bottomrule
\end{tabular}
\end{table}

\section{Maximum-likelihood consensus pose}
\label{app:mle_pose}
To establish a single robust 3D joint estimate per timestep from $N$ overlapping views, we form an MLE-weighted consensus pose. Let $J_{c}^{(i)} \in \mathbb{R}^3$ represent a joint position estimated by the monocular model from camera $i$ in its local coordinate frame. We model the per-camera measurement noise as a Gaussian distribution centered at the true camera-frame joint position. To account for the fact that depth is the dominant error mode in monocular reconstruction, the covariance matrix $\Sigma_c^{(i)} \in \mathbb{R}^{3 \times 3}$ is elongated along that camera's depth axis.

Before fusing the estimates, we must lift each camera's prediction into the shared world coordinate system (the tennis court frame). Given the camera-to-world calibration parameters $R_{\text{calib\_c2w}}^{(i)} \in \mathbb{R}^{3 \times 3}$ and $T_{\text{calib\_c2w}}^{(i)} \in \mathbb{R}^3$, the lifted world-frame joint observation is:
\begin{equation}
    J_{w}^{(i)} = R_{\text{calib\_c2w}}^{(i)} J_{c}^{(i)} + T_{\text{calib\_c2w}}^{(i)}
\end{equation}

To ensure correct coordinate transformation, the noise covariance matrix must also be transformed into the world frame. The world-frame covariance $\Sigma_w^{(i)}$ is computed as:
\begin{equation}
    \Sigma_w^{(i)} = R_{\text{calib\_c2w}}^{(i)} \Sigma_c^{(i)} \left(R_{\text{calib\_c2w}}^{(i)}\right)^T
\end{equation}

Let the true consensus world-frame joint position be $P$. Assuming the measurement errors across the $N$ cameras are independent, the joint likelihood of observing the set of lifted estimates $\mathcal{D} = \{J_{w}^{(1)}, \dots, J_{w}^{(N)}\}$ is the product of the individual Gaussian likelihoods:
\begin{equation}
    L(P | \mathcal{D}) = \prod_{i=1}^{N} \frac{1}{\sqrt{(2\pi)^3 |\Sigma_w^{(i)}|}} \exp\left(-\frac{1}{2} (J_{w}^{(i)} - P)^T (\Sigma_w^{(i)})^{-1} (J_{w}^{(i)} - P)\right)
\end{equation}

The Maximum Likelihood Estimate (MLE) for the true pose, $P_{\text{MLE}}$, is found by minimizing the negative log-likelihood:
\begin{equation}
    -\ln L(P | \mathcal{D}) = \frac{1}{2} \sum_{i=1}^{N} (J_{w}^{(i)} - P)^T (\Sigma_w^{(i)})^{-1} (J_{w}^{(i)} - P) + C
\end{equation}

Taking the derivative with respect to $P$ and setting it to zero yields:
\begin{equation}
    \frac{\partial (-\ln L)}{\partial P} = - \sum_{i=1}^{N} (\Sigma_w^{(i)})^{-1} (J_{w}^{(i)} - P) = 0
\end{equation}

Solving for $P$ yields the closed-form solution, which is a precision-weighted average of the lifted per-view estimates:
\begin{align}
    \left( \sum_{i=1}^{N} (\Sigma_w^{(i)})^{-1} \right) P &= \sum_{i=1}^{N} (\Sigma_w^{(i)})^{-1} J_{w}^{(i)} \\
    P_{\text{MLE}} &= \left( \sum_{i=1}^{N} (\Sigma_w^{(i)})^{-1} \right)^{-1} \sum_{i=1}^{N} (\Sigma_w^{(i)})^{-1} J_{w}^{(i)}
\end{align}

This formulation ensures that the highly uncertain depth axes from individual monocular cameras are appropriately down-weighted during spatial fusion. We use the MLE pose for training on in-the-wild videos of tennis practices and matches, as shown in \cref{fig:mle_vis_pose}.

\subsection{Human pose reconstruction from video}
\label{app:hpe_recon}
Following PromptHMR\cite{wang2025prompthmrpromptablehumanmesh}, we first detect people and 2D joint poses using ViTPose~\cite{xu2022vitposesimplevisiontransformer}, and associate people between frames by propagating masks and bounding boxes with SAM2~\cite{ravi2024sam2segmentimages}. We estimate camera intrinsics and extrinsics using Droid-SLAM~\cite{teed2022droidslamdeepvisualslam} and ZoeDepth~\cite{bhat2023zoedepthzeroshottransfercombining}. Next, we recover 3D human pose estimates with PromptHMR~\cite{wang2025prompthmrpromptablehumanmesh}, producing per-frame pose estimates of all $p$ detected people in the scene with 3D SMPL-X~\cite{pavlakos2019smplx} parameters: $H = \smash{\{(d_t, \phi_t, \beta_t, \theta_t)\}_{t=0}^{T}}$, with translation $d_t \in \mathbb{R}^{p \times 3}$, body pose $\theta_t \in \mathbb{R}^{p \times 21 \times 3}$, orientation $\phi_t \in \mathbb{R}^{p \times 3}$, and shape $\beta_t \in \mathbb{R}^{p \times 10}$.

\section{RL Training Details}
\label{app:training}
We set our training hyperparameters according to the standard of the field for ease of reproducibility. We train using the Asymmetric Actor Critic PPO algorithm \cite{schulman2017proximal}. We use a discount factor of $\gamma = 0.99$, a GAE of $\lambda =0.95$, learning rate of $1e-3$ and desired KL of $0.01$. We set the max gradient norm to $1.0$ and the entropy coefficient to $0.005$, use $5$ learning epochs per rollout, and each rollout is of length $24$. We train on a single NVIDIA 4090 GPU with 4096 environments. Finally, both our actor and critic networks have three hidden layers of $\left[512, 256, 128\right]$. 

\subsection{Observation Details}
\label{app:observations}

  \newcolumntype{C}{>{\centering\arraybackslash}X}
  \begin{table}[h]
  \label{tab:observations}
  \mediumpfont
  \caption{
      \textbf{Observation space.}
      Observations used by the actor (policy) and critic (value function).
      Noise is injected only for actor observations during training.
      All positions and orientations in the robot base frame are denoted with subscript $b$.
  }

  \centering
  \setlength{\tabcolsep}{3.5pt}

  \resizebox{\textwidth}{!}{%
  \begin{tabular}{llccc @{}}

  \toprule

  \textbf{Observation}
   & \textbf{Description}
   & \makecell{Actor\\(policy)}
   & \makecell{Critic\\(value fn.)}
   & \makecell{Noise\\$\mathcal{U}(\pm\,\cdot)$}
   \\

  \midrule

  Motion command
   & Reference Body Positions, target positions, velocities, orientations
   & \greencheck & \greencheck & ---
   \\

  Anchor orientation ${R}^{\mathrm{anc}}_{b}$
   & Rotation of anchor body in robot frame
   & \greencheck & \greencheck & $0.05$
   \\

  Base angular velocity ${\omega}$
   & IMU angular velocity
   & \greencheck & \greencheck & $0.20$
   \\

  Joint positions ${q}$
   & Joint angles relative to default pose
   & \greencheck & \greencheck & $0.01$
   \\

  Joint velocities $\dot{{q}}$
   & Joint velocities relative to default
   & \greencheck & \greencheck & $0.50$
   \\

  Last action ${a}_{t-1}$
   & Previous joint-position targets
   & \greencheck & \greencheck & ---
   \\

  \midrule

  Anchor position ${p}^{\mathrm{anc}}_{b}$
   & Position of anchor body in robot frame
   & \redcross & \greencheck & ---
   \\

  Body positions ${p}^{\mathrm{body}}_{b}$
   & Positions of all tracked bodies in robot frame
   & \redcross & \greencheck & ---
   \\

  Body orientations ${R}^{\mathrm{body}}_{b}$
   & Rotations of all tracked bodies in robot frame
   & \redcross & \greencheck & ---
   \\

  Base linear velocity ${v}$
   & IMU linear velocity
   & \redcross & \greencheck & ---
   \\

  \bottomrule
  \end{tabular}
  }%
\end{table}

\subsection{Action Details}
\label{app:actions}

We use unbounded actions without activations on the actor output following the style of \cite{rudin2022learning}. We do not clip our actions, instead taking the output of the actor, sending it through an action scaling to allow the network outputs to live in $[-1, 1]$, add this value to a default joint position, then send these as desired joint positions. PD controllers at each joint track the positions sent by the actor network at 200hz, with gains from \cite{liao2025beyondmimic}.  

\subsection{Reward Details}
\label{app:rewards}
\newcolumntype{C}{>{\centering\arraybackslash}X}
  \begin{table}[h]
  \label{tab:rewards}
  \mediumpfont
  \caption{
      \textbf{Reward terms.}
      $\hat{\cdot}$ denotes quantities from the reference motion clip.
      Subscript $b$ denotes the robot base frame.
      $N$ is the number of tracked bodies.
      $\mathrm{Log}:\mathrm{SO(3)}\!\to\!\mathbb{R}^3$ is the rotation-vector logarithm.
      Target rewards are gated to the configured phase window.
  }

  \centering
  \setlength{\tabcolsep}{3.5pt}

  \resizebox{\textwidth}{!}{%
  \begin{tabular}{llccc @{}}

  \toprule

  \textbf{Reward}
   & \textbf{Formula}
   & \textbf{Description}
   & \textbf{Weight} $w$
   & $\sigma$
   \\

  \midrule
  \multicolumn{5}{l}{\textit{Motion imitation}} \\
  \midrule

  Anchor position
   & $\exp\!\bigl(-\|{p}^{\mathrm{anc}} - \hat{{p}}^{\mathrm{anc}}\|^2/\sigma^2\bigr)$
   & Global anchor body position error
   & $0.0$
   & $0.30$
   \\

  Anchor orientation
   & $\exp\!\bigl(-\|\mathrm{Log}(\hat{{R}}^{\mathrm{anc}\top}{R}^{\mathrm{anc}})\|^2/\sigma^2\bigr)$
   & Global anchor body orientation error
   & $0.5$
   & $0.40$
   \\

  Body positions
   & $\exp\!\bigl(-\tfrac{1}{N}\sum_i\|{p}^i_b - \hat{{p}}^i_b\|^2/\sigma^2\bigr)$
   & Mean relative body position error in base frame
   & $1.0$
   & $0.30$
   \\

  Body orientations
   & $\exp\!\bigl(-\tfrac{1}{N}\sum_i\|\mathrm{Log}(\hat{{R}}^{i\top}_b{R}^i_b)\|^2/\sigma^2\bigr)$
   & Mean relative body orientation error in base frame
   & $1.0$
   & $0.40$
   \\

  Body linear velocities
   & $\exp\!\bigl(-\tfrac{1}{N}\sum_i\|{v}^i - \hat{{v}}^i\|^2/\sigma^2\bigr)$
   & Mean global body linear velocity error
   & $1.0$
   & $1.00$
   \\

  Body angular velocities
   & $\exp\!\bigl(-\tfrac{1}{N}\sum_i\|{\omega}^i - \hat{{\omega}}^i\|^2/\sigma^2\bigr)$
   & Mean global body angular velocity error
   & $1.0$
   & $\pi$
   \\

  \midrule
  \multicolumn{5}{l}{\textit{Target achievement}} \\
  \midrule

  Target position
   & $\exp(-||p^* - p||^2/{\sigma^2})\mathbbm{1}_{\tau \in \Omega}$
   & End-effector position error at target phase
   & $1.0$
   & $0.30$
   \\

  Target orientation
   & $\exp(-(1-\langle n^*, n\rangle)/{\sigma^2})\mathbbm{1}_{\tau \in \Omega}$
   & End-effector primary-axis alignment error at target phase
   & $1.0$
   & $1.00$
   \\

  Target velocity
   & $\exp(-||\nu^* - \nu||^2/{\sigma^2})\mathbbm{1}_{\tau \in \Omega}$
   & End-effector velocity error at target phase
   & $1.0$
   & $1.00$
   \\

  \midrule
  \multicolumn{5}{l}{\textit{Regularization}} \\
  \midrule

  Action rate
   & $-\|{a}_t - {a}_{t-1}\|^2$
   & Penalizes high-input actions
   & $-0.10$
   & ---
   \\

  Joint limits
   & $-\sum_i \max(q_i - q_i^{\max},\, 0) + \max(q_i^{\min} - q_i,\, 0)$
   & Penalizes joint position limit violations
   & $-10.0$
   & ---
   \\

  Self-collisions
   & $-\sum_c F_c\;{1}[F_c > F_{\mathrm{thr}}]$
   & Penalizes self-contact forces above $F_{\mathrm{thr}}{=}10\,\mathrm{N}$
   & $-10.0$
   & ---
   \\

  \bottomrule
  \end{tabular}
  }%
  \end{table}

\subsection{Termination Details}
\label{app:terminations}

Our training environments terminate early as per the recommendation of \cite{peng2018deepmimic} if any of the following criteria are met. We track the error of many of the robot's bodies Z coordinate, terminating if any fall below 0.25m below the reference. We do not track any of the other directions, as for imperfect translations in the motion or if we want to move to a slightly different target, deviation from the anchor position in the lateral directions is desired. We terminate if the anchor body's orientation exceeds 0.8 radians, ensuring the robot stays upright. Finally we use an episode length of 10s, so at this point the episode times out.

\subsection{Ablations}
\label{app:ablations}

For our ablations, we measure four metrics. Training stability is measured in iterations to convergence, defined as the point at which the target position reward hits 95\% of its final value consistently, rounded to the nearest 500 iterations (n/a indicates that it never converged). Note that each iteration, run on a single RTX4090 GPU, takes about an average of 1.15 seconds. Success rate is measured as rate at which the robot succeeds at the task without falling or self collision, average targeting error is the minimum distance between the target point and the interfacing link over the motion, and average phase error is the average time in seconds between the desired and actual contact with the target point. Each of the evaluations are run on 4096 parallel environments with randomized targets within 2m laterally, 0.5m vertically, and 0.2m longitudinally of the robot's anchor point. Averages and standard deviations are taken across environments. 

\newcolumntype{C}{>{\centering\arraybackslash}X}
\begin{table}[h]
\label{tab:your_table_label}
\mediumpfont
\caption{
    \textbf{Target Phase Length + Reward Density} 
    }

\centering
\setlength{\tabcolsep}{3.5pt} 

\resizebox{\textwidth}{!}{%
\begin{tabular}{lccccc @{}}

\toprule
 & \makecell{\# Frames \\ Of Contact}
 & \makecell{Convergence Iterate}
 & \makecell{Success Rate}
 & \makecell{Target Position Error (m)}
 & \makecell{Target Phase Error (s)}

 \\

\midrule
\rowcolor{gray!15}

\textbf{}
\textbf{}
    & 1
    & 3000
    & 0.9939
    & 0.0205 {\tiny (0.0108)}
    & 0.1665 {\tiny (0.2191)}
    \\
\textbf{}
    & 3
    & 3000
    & 0.9768
    & 0.0499 {\tiny (0.0537)}
    & 0.1346 {\tiny (0.2462)}
    \\

\textbf{}
    & 5
    & 4000
    & 0.9846
    & 0.0522 {\tiny (0.0548)}
    & 0.1343 {\tiny (0.2149)}
    \\

\textbf{}
    & 7
    & 5500
    & 0.9854
    & 0.0486 {\tiny (0.0552)}
    & 0.1429 {\tiny (0.2467)}
    \\

\textbf{}
    & 15
    & n/a
    & 0.9895
    & 0.0927 {\tiny (0.0961)}
    & 0.2338 {\tiny (0.3822)}
    \\

\textbf{}
    & 30
    & n/a
    & 1.0000
    & 0.3338 {\tiny (0.1780)}
    & 0.4508 {\tiny (0.7283)}
    \\

\textbf{}
    & 200
    & n/a
    & 0.9985
    & 0.3253 {\tiny (0.1771)}
    & 0.5057 {\tiny (0.8225)}
    \\

\bottomrule
\end{tabular}
}%
\end{table}

\begin{figure}[h]
  \centering
  \includegraphics[width=\linewidth]{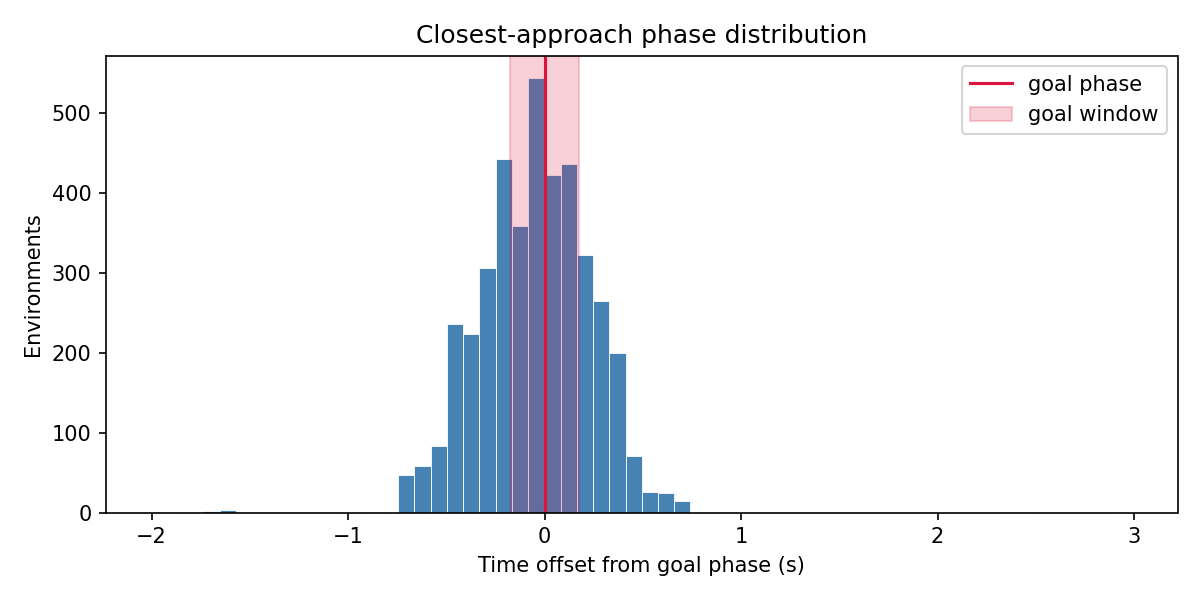}
  \caption{
  \textbf{\method Phase Histogram.}
  We get density of target hit time even with very sparse rewards. We define the goal window as the phase range that we emperically determine is sufficient to get a solid hit. 
  }
  \label{fig:phase_histogram}
\end{figure}

We first ablate the phase density. Our nominal setup concentrates the entire goal reward at the impact frame; spreading the same reward over a longer phase window monotonically degrades targeting, and at the limit (200 frames) the policy collapses to pure imitation without hitting the target. Despite the sparsity of the single-frame reward, Figure~\ref{fig:phase_histogram} shows dense behavior around the target phase — a favorable indicator of robustness in hit time.
\newcolumntype{C}{>{\centering\arraybackslash}X}
\begin{table}[h]
\label{tab:mean_variation}
\mediumpfont
\caption{
    \textbf{Target Mean Variation} 
    }

\centering
\setlength{\tabcolsep}{3.5pt} 

\resizebox{\textwidth}{!}{%
\begin{tabular}{lcccc @{}}

\toprule

 \makecell{$||\bar{p}-p^*||$}
 & \makecell{Convergence Iterate}
 & \makecell{Success Rate}
 & \makecell{Target Position Error (m)}
 & \makecell{Target Phase Error (s)}

 \\

\midrule
\rowcolor{gray!15}

\textbf{0.0}
    & 3000
    & 0.9939
    & 0.0205 {\tiny (0.0108)}
    & 0.1665 {\tiny (0.2191)}
    \\

\textbf{0.374}
    & 13000
    & 0.9907
    & 0.0523 {\tiny (0.0553)}
    & 0.1302 {\tiny (0.2334)}
    \\

\textbf{0.412}
    & 17000
    & 0.9824
    & 0.0537 {\tiny (0.0524)}
    & 0.1278 {\tiny (0.2147)}
    \\

\textbf{0.469}
    & 22000
    & 0.9819
    & 0.0754 {\tiny (0.0890)}
    & 0.1500 {\tiny (0.3325)}
    \\

\textbf{0.538}
    & n/a
    & 0.9841
    & 0.0925 {\tiny (0.1024)}
    & 0.1643 {\tiny (0.3645)}
    \\

\bottomrule
\end{tabular}
}%
\end{table}

\newcolumntype{C}{>{\centering\arraybackslash}X}
\begin{table}[h]
\label{tab:target_std_dev}
\mediumpfont
\caption{
    \textbf{Target Standard Deviation Variation} 
    }

\centering
\setlength{\tabcolsep}{3.5pt} 

\resizebox{\textwidth}{!}{%
\begin{tabular}{lcccc @{}}

\toprule

 & \makecell{Convergence Iterate}
 & \makecell{Success Rate}
 & \makecell{Target Position Error (m)}
 & \makecell{Target Phase Error (s)}

 \\

\midrule
\multicolumn{5}{l}{\small \textbf{$\sigma^2_y, \sigma^2_z$} (within hitting plane)} \\
\midrule
\rowcolor{gray!15}
\textbf{0.20}
    & 3000
    & 0.9939
    & 0.0205 {\tiny (0.0108)}
    & 0.1665 {\tiny (0.2191)}
    \\

\textbf{0.25}
    & 6500
    & 0.9731
    & 0.0465 {\tiny (0.0489)}
    & 0.1376 {\tiny (0.2402)}
    \\

\textbf{0.30}
    & 7000
    & 0.9893
    & 0.044 {\tiny (0.048)}
    & 0.1299 {\tiny (0.1948)}
    \\

\textbf{0.35}
    & 9500
    & 0.9946
    & 0.0428 {\tiny (0.0517)}
    & 0.137 {\tiny (0.2373)}
    \\

\textbf{0.40}
    & 9000
    & 0.9988
    & 0.0406 {\tiny (0.0503)}
    & 0.1321 {\tiny (0.2169)}
    \\

\textbf{1.0}
    & n/a
    & 1.0000
    & 0.3569 {\tiny (0.1831)}
    & 0.467 {\tiny (0.7425)}
    \\

\midrule
\multicolumn{5}{l}{\small \textbf{$\sigma^2_x$} (out of hitting plane)} \\
\midrule

\textbf{0.05}
    & 5000
    & 0.9507
    & 0.0465 {\tiny (0.052)}
    & 0.1317 {\tiny (0.2194)}
    \\
\rowcolor{gray!15}
\textbf{0.10}
    & 3000
    & 0.9939
    & 0.0205 {\tiny (0.0108)}
    & 0.1665 {\tiny (0.2191)}
    \\

\textbf{0.15}
    & 4000
    & 0.958
    & 0.0458 {\tiny (0.0539)}
    & 0.1307 {\tiny (0.2278)}
    \\

\textbf{0.20}
    & 5000
    & 0.9722
    & 0.0474 {\tiny (0.0533)}
    & 0.1343 {\tiny (0.2330)}
    \\

\textbf{0.25}
    & 6000
    & 0.9768
    & 0.0495 {\tiny (0.0515)}
    & 0.1318 {\tiny (0.2059)}
    \\

\textbf{1.0}
    & 18000
    & 0.9727
    & 0.0536 {\tiny (0.0561)}
    & 0.1333 {\tiny (0.2352)}
    \\

\bottomrule
\end{tabular}}
\end{table}

\newcolumntype{C}{>{\centering\arraybackslash}X}
\begin{table}[h]
\label{tab:motion_retiming}
\mediumpfont
\caption{
    \textbf{Motion Retiming} 
    }

\centering
\setlength{\tabcolsep}{3.5pt} 

\resizebox{\textwidth}{!}{%
\begin{tabular}{lcccc @{}}

\toprule

 & \makecell{Convergence Iterate}
 & \makecell{Success Rate}
 & \makecell{Target Position Error (m)}
 & \makecell{Target Phase Error (s)}

 \\

\midrule
\multicolumn{5}{l}{{Eval Hz = Train Hz}} \\
\midrule
\textbf{0.33}
    & 7000
    & 0.9971
    & 0.1820 {\tiny (0.1714)}
    & 0.4739 {\tiny (1.2329)}
    \\

\textbf{0.66}
    & 6000
    & 0.9988
    & 0.0383 {\tiny (0.0485)}
    & 0.1853 {\tiny (0.3047)}
    \\
\rowcolor{gray!15}
\textbf{1.0}
    & 3000
    & 0.9939
    & 0.0205 {\tiny (0.0108)}
    & 0.1665 {\tiny (0.2191)}
    \\

\textbf{1.33}
    & 8500
    & 0.9963
    & 0.0344 {\tiny (0.0253)}
    & 0.1054 {\tiny (0.1485)}
    \\

\textbf{1.66}
    & 9000
    & 0.9968
    & 0.0346 {\tiny (0.0261)}
    & 0.0902 {\tiny (0.1203)}
    \\

\midrule
\multicolumn{5}{l}{{Eval Hz = 1.0}} \\
\midrule

\textbf{0.33}
    & 7000
    & 0.7793
    & 0.2384 {\tiny (0.1227)}
    & 0.3069 {\tiny (0.5924)}
    \\

\textbf{0.66}
    & 6000
    & 0.8582
    & 0.0754 {\tiny (0.0518)}
    & 0.1383 {\tiny (0.2321)}
    \\
\rowcolor{gray!15}
\textbf{1.0}
    & 3000
    & 0.9939
    & 0.0205 {\tiny (0.0108)}
    & 0.1665 {\tiny (0.2191)}
    \\

\textbf{1.33}
    & 8500
    & 0.9634
    & 0.0661 {\tiny (0.0486)}
    & 0.1326 {\tiny (0.2238)}
    \\

\textbf{1.66}
    & 9000
    & 0.8318
    & 0.1023 {\tiny (0.0541)}
    & 0.1757 {\tiny (0.2924)}
    \\

\bottomrule
\end{tabular}
}%
\end{table}

\newcolumntype{C}{>{\centering\arraybackslash}X}
\begin{table}[h]

\mediumpfont
\caption{
    \textbf{Motion Count} 
}
\label{app:motion_count}

\centering
\setlength{\tabcolsep}{3.5pt} 

\resizebox{\textwidth}{!}{%
\begin{tabular}{lcccc @{}}

\toprule

 & \makecell{Training Stability}
 & \makecell{Success Rate}
 & \makecell{Target Position Error (m)}
 & \makecell{Target Phase Error (s)}

 \\

\midrule

\textbf{10}
    & \phantom{0}7000& 0.9878 
    & 0.0319 {\tiny (0.0158)}
    & 0.5750 {\tiny (0.2834)}
    \\

\textbf{15}
    & \phantom{0}4700& 0.9944 
    & 0.0346 {\tiny (0.0265)}
    & 0.5579 {\tiny (0.8532)}

    \\

\textbf{20}
    & 19800& 0.9277 
    & 0.0413 {\tiny (0.0331)}
    & 0.7102 {\tiny (1.0616)}

    \\

\textbf{23}
    & \phantom{0}9500& 0.9265 
    & 0.0422 {\tiny (0.0327)}
    & 0.7189 {\tiny (1.0247)}

    \\
\textbf{28}
    & 17900& 0.969 
    & 0.0453 {\tiny (0.0332)}
    & 0.5153 {\tiny (0.8635)}

    \\
\textbf{34}
    & \phantom{0}7200& 0.9082 
    & 0.0495 {\tiny (0.0386)}
    & 0.5077 {\tiny (0.8005)}
    \\

\bottomrule
\end{tabular}
}%
\end{table}

We ablate the command formulation along three axes. Varying the mean offset from the nominal target, the policy tracks reliably out to roughly 0.6m on tennis tasks, beyond which training stability degrades as the policy takes longer to discover that hitting the target — not just imitating the motion — earns reward. Varying the target standard deviation leaves average targeting error roughly constant until the variance becomes very large, at which point the policy reverts to pure imitation. Finally, varying motion speed in training and deployment shows that kinematic retargeting combined with our phase observation yields robustness to retiming: the policy is best when training and deployment speeds match, but remains reasonably performant when they do not.

\subsection{Simulation Parameters and Domain Randomization}
\label{app:simulation}

We run Mjlab \cite{zakka2026mjlab} at 200hz with a control decimation of 4, giving a policy a frequency of 50hz. We perform identical domain randomization to the Mjlab imitation policy setup, randomizing pushes on our robot, our base COM offset, encoder bias, and foot friction. 

\section{Hardware Results}
\label{app:hardware_results}

We evaluate on three hardware tasks. Tennis success is a racket–ball contact; soccer success is a foot–ball contact; box pick-and-place success requires lifting a 0.3m x 0.3m x 0.5m box and dropping it within 1m of the goal. Each task has a slow variant (stationary, gently tossed, or swung on a rope) and a fast variant (balls thrown or kicked at 4–8 m/s), with objects placed up to 3m laterally for sports and within a 3m semicircle in front of the robot for boxes. We run 20 trials per condition. The robot achieves a perfect hit rate on slow sports targets, with box pickup somewhat lower: every box failure was an imperfect grasp, unsurprising given that we do not train recovery behaviors. Performance on fast targets degrades gracefully with ball speed, and every failure traced to the perception stack; the tennis–soccer gap reflects the difficulty of tracking a bouncing tennis ball on uneven floor. Notably, the robot tolerated pushes and varied room and lighting conditions, and never fell even on missed hits, which are properties we attribute to the abstraction. Training with the estimator in the loop, or with motion-capture failures in domain randomization, should close the remaining gap; we leave this to future work.

\subsection{Estimation Details}
\label{app:estimation}

We find the target and the robot using an OptiTrack Motion Capture setup running at 120hz composed of 8 cameras capturing an effective 4.5m by 7.5m by 3m volume.



\end{document}